\documentclass[review]{elsarticle}
\makeatletter
\def\ps@pprintTitle{%
 \let\@oddhead\@empty
 \let\@evenhead\@empty
 \def\@oddfoot{\centerline{\thepage}}%
 \let\@evenfoot\@oddfoot}
\makeatother
 
\usepackage{lineno}
\usepackage{wrapfig}
\usepackage{color}
\usepackage{times}
\usepackage{epsfig}
\usepackage{graphicx}
\usepackage{amsmath}
\usepackage{amsthm}
\usepackage{mathrsfs} 
\usepackage{amssymb}
\def\R{\mathbb{R}}

\def\L{{\cal L}}
\def\MP{{\operatorname{MP}}}
\def\R{{\mathbb R}}
\newcommand{\etal}{\textit{et al.}}
\usepackage{wrapfig}
\usepackage{subfig}

\modulolinenumbers[5]

\def\R{\mathbb{R}}
\def\L{\mathcal{L}}
\def\O{\mathcal{O}}

\begin{document}

\begin{frontmatter}

\title{Discovery and visualization of structural biomarkers from MRI using transport-based morphometry}

\author{Shinjini Kundu, Soheil Kolouri, Kirk I. Erickson, Arthur F. Kramer, Edward McAuley, Gustavo K. Rohde}

\begin{abstract}
Disease in the brain is often associated with subtle, spatially diffuse, or complex tissue changes that may lie beneath the level of gross visual inspection, even on magnetic resonance imaging (MRI). Unfortunately, current computer-assisted approaches that examine pre-specified features, whether anatomically-defined (i.e. thalamic volume, cortical thickness) or based on pixelwise comparison (i.e. deformation-based methods), are prone to missing a vast array of physical changes that are not well-encapsulated by these metrics. In this paper, we have developed a technique for automated pattern analysis that can fully determine the relationship between brain structure and observable phenotype without requiring any \textit{a priori} features. Our technique, called transport-based morphometry (TBM), is an image transformation that maps brain images losslessly to a domain where they become much more separable. The new approach is validated on structural brain images of healthy older adult subjects where even linear models for discrimination, regression, and blind source separation enable TBM to independently discover the characteristic changes of aging and highlight potential mechanisms by which aerobic fitness may mediate brain health later in life. TBM is a generative approach that can provide visualization of physically meaningful shifts in tissue distribution through inverse transformation. The proposed framework is a powerful technique that can potentially elucidate genotype-structural-behavioral associations in myriad diseases. 

\end{abstract}

\begin{keyword}
\texttt{magnetic resonance imaging, computer-aided detection, aging, transport-based morphometry}
\end{keyword}

\end{frontmatter}

\linenumbers

\section{Introduction}

\begin{figure}
\includegraphics[width = 0.8\columnwidth]{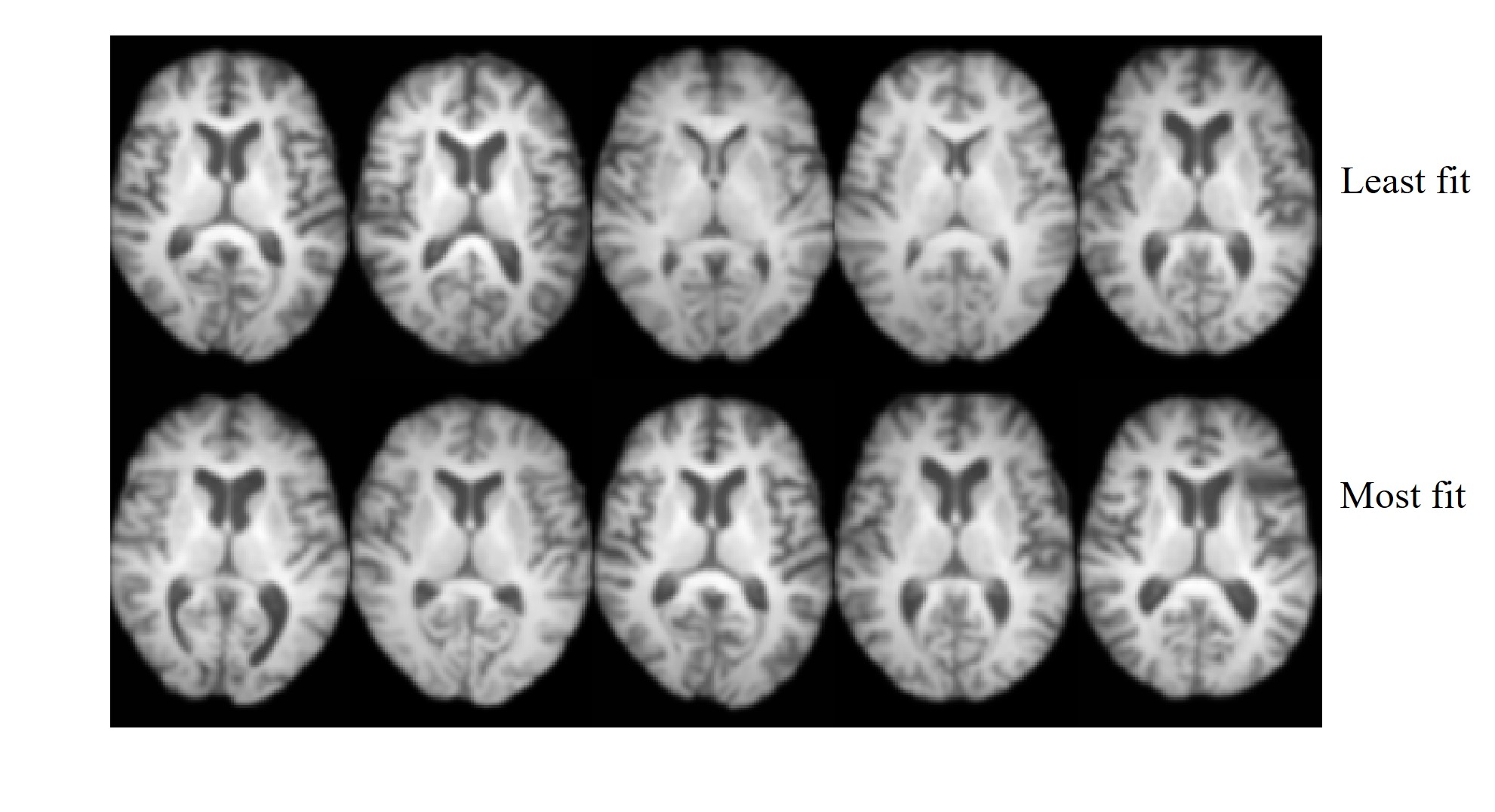}
\centering
\vspace{-1.5em}
\caption{\footnotesize{MR images belonging to 10 older adults in their $6th$ or $7th$ decades of life. The images correspond to subjects who are either $2\sigma$ above or below the mean aerobic fitness as assessed by $vO_2$ L/min. The goal is to determine whether there is a common morphologic feature that separates these groups, and if so, visualize it in a physically interpretable manner.}}
\label{fig:unfit}
\end{figure}

Recent advances in magnetic resonance imaging (MRI) technology have enabled high-resolution imaging across many new modalities. Tissue properties can be now be measured at an unprecedented level of precision and detail. These developments hold promise to illuminate structural changes underlying diseases commonly considered medical mysteries. Unfortunately, the changes can often be subtle, spatially diffuse, and complex, escaping detection by visual inspection. For example, Figure \ref{fig:unfit} demonstrates how the common morphologic pattern that differentiates individuals who are most aerobically fit versus those who are least fit {defies identification by gross inspection alone}. Thus, there is a growing role for computer-aided techniques to aid in vision and detection of morphologic patterns from MRI. Computer-aided techniques are needed to answer the following questions: are there morphologic differences that differentiate these groups? If so, what are they?

\begin{figure}
\begin{centering}
\includegraphics[width = 0.8\columnwidth]{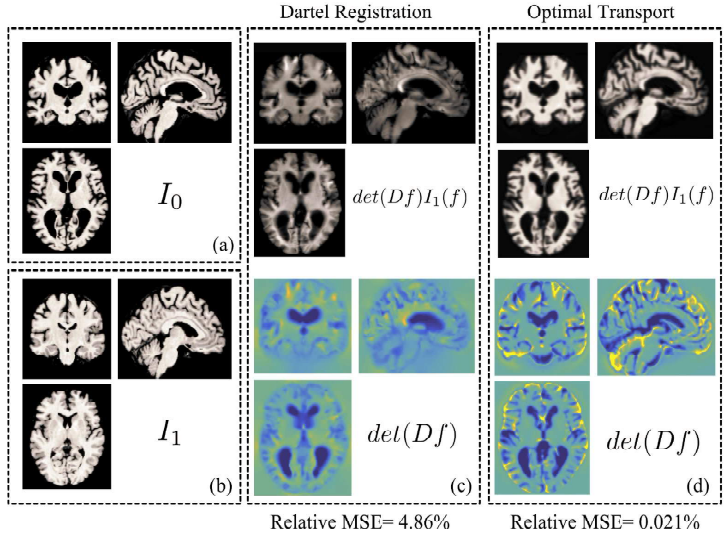}
\vspace{-0.5em}
\caption{\footnotesize{Compared to deformation fields computed using DARTEL \cite{ashburner2007fast}, transport maps computed using optimal mass transport (OMT) captures both shape and texture differences between $I_0$ and $I_1$ and match images perfectly, up to an interpolation error. The three boxes in the top row should all look the same. However, deformation fields lose texture variation information, thus resulting in high MSE when attempting to match source and target images. \label{fig:dartel}}}
\end{centering}
\end{figure}

\begin{figure}
\begin{centering}
\includegraphics[width = 0.8\columnwidth]{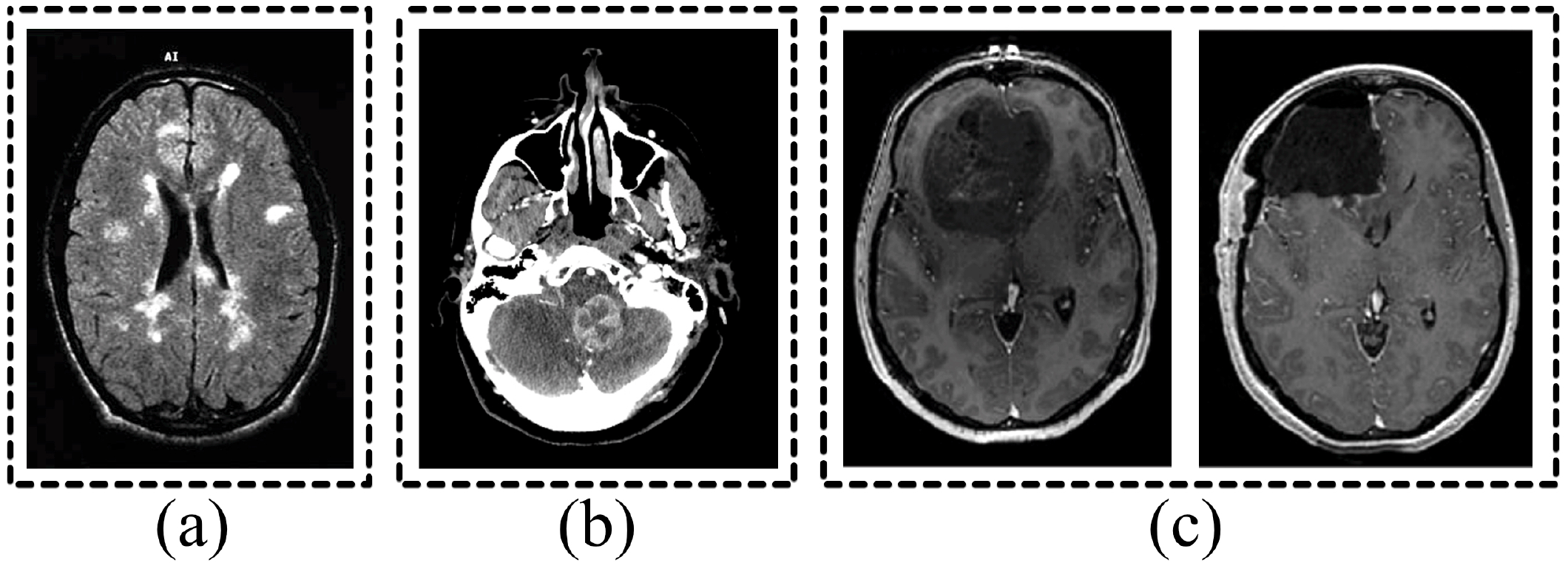}
\vspace{-1em}
\caption{\footnotesize{Neurologic conditions where pathology affects biophysical properties of tissue manifesting as texture variation. (a) Multiple sclerosis (source: \cite{jeffery2002use}), (b) metastatic breast cancer tumor (source: Dept. of Radiology, University of Pittsburgh Medical Center), (c) GBM (left) and debulking procedure (right) (source: \cite{Carlson2011})
}}
\vspace{-1.5em}
\label{fig:examples}
\end{centering}
\end{figure}

Unfortunately, traditional techniques for MRI analysis have difficulty with analysis in the image domain as well, as they require features to be pre-specified and are prone to missing a multitude of physical changes that are not adequately assessed by these finite feature sets. For example, popular biomedical image analysis softwares such as WND-CHRM \cite{shamir2008wndchrm} or FreeSurfer \cite{fischl2000measuring} extract a number of pre-specified numerical descriptors, such thickness, volume, texture statistics, etc., from the images and test whether these quantities are statistically different between image sets using a trial and error approach. In fact, WND-CHRM extracts nearly 3000 generic features from the images for testing. However, not only is testing descriptors a tedious process, numerical descriptors such as SIFT, Gabor features, or histogram statistics often do not have direct biological meaning. Another major limitation of these approaches is that the analysis does not incorporate anatomic prior information, missing an opportunity to compare variations in terms of known anatomy. Deformation-based methods, which include deformation-ased morphometry (DBM) \cite{ashburner1998identifying}, tensor-based morphometry \cite{ashburner2000tensor}, and voxel-based morphometry (VBM) \cite{ashburner2000voxel}, also have limitations. Deformation-based methods rely on a nonrigid registration to align images before comparing them pixelwise. In practice, perfect structural and functional alignment cannot be ensured, and subtle changes in pixel alignment can vastly change the results obtained and undermine accuracy \cite{bookstein2001voxel}. Another limitation is these techniques is that the deformation fields are not unique. Hence, the results of Tensor-Based Morphometry and deformation-based morphometry, which compare pixelwise across the determinant of Jacobian or deformation fields respectively, will vary depending on the particular field generated by the algorithm. While these methods incorporate anatomical information, projecting images onto individual pixels assumes that the changes are localized into clusters and misses spatially diffuse changes. Furthermore, deformation fields model changes in local volume expansion/contraction in order to match gross shapes. However, as Figure \ref{fig:dartel} illustrates, deformation fields cannot fully match images with zero error because they cannot quantify differences in tissue topology or texture. Figure \ref{fig:examples} illustrates several neurologic diseases for which the main variation is in tissue texture rather than brain volume contraction/expansion - multiple sclerosis lesions and brain tumors. The authors of deformation-based techniques state that these registration-based methods are a way to index into pixels based on amount of gray matter per unit volume \cite{salmond2002distributional}, but cannot offer insight into the physical meaning of these changes \cite{friston2004generative} as these methods are not \textit{generative}. If a technique is not generative, then an observable datapoint, or a brain MR image cannot be generated given a feature set like a density map. However, a generative method would afford the ability to visualize the shifts in morphologic profile as dynamic changes across a series of brain MRIs to illuminate structural mechanisms.


In this paper, we describe transport-based morphometry (TBM) \cite{wang2013linear, wang2011optimal, basu2014detecting, park2017cumulative, kolouri2016continuous}, which has the potential to enable fully automated MRI analysis without loss of information. Rather than analyzing images in the image domain where they may not be easily separable, we first transform them to a domain that enhances separability. Prior work demonstrates that when we transform 1D and 2D signals in other applications using TBM \cite{wang2013linear, wang2011optimal, basu2014detecting, park2017cumulative, kolouri2016continuous, kolouri2016radon}, complex and nonlinear morphology in the image domain can be described by linear classification and regression models in the transform domain. Furthermore, the key advance of TBM is that it is generative and enables direct visualization of the interface between signal classes through inverse TBM transformation. The TBM technique computes the distance needed to morph one image with respect to a common template using the mathematics of optimal mass transport (OMT). As Figure \ref{fig:dartel} illustrates, unlike deformation-based approaches, OMT can match both shape and texture variations simultaneously; thus, information is not missed. However, TBM has never been applied for MRI-based detection and pattern analysis as current formulations and solutions to TBM are designed for smaller signals \cite{wang2013linear, wang2011optimal, basu2014detecting, park2017cumulative, kolouri2016continuous, kolouri2016radon}. 

In this paper, we demonstrate a TBM framework that is suitable for analysis of radiology data, the majority of which comprises three-dimensional data. We hypothesize that transforming MRI data using the new TBM approach can facilitate both discovery as well as visualization of discriminating differences in a manner similar to 1D and 2D signal analysis previously reported \cite{wang2013linear, wang2011optimal, basu2014detecting, park2017cumulative, kolouri2016continuous, kolouri2016radon}. Ultimately, if TBM could assess structural changes underlying clinical phenotypes in a fully automated manner without losing information, and in addition, visualize the shifts in tissue distribution as a series of radiology images, it could represent a breakthrough for scientific understanding as well as identification of objective clinical markers.



The specific contributions of this work are: 

\begin{itemize}
\item{Novel, robust formulation and solution of transport-based morphometry (TBM) enabling its first application to radiology data and computational validation}
\item{A new image transform to facilitate pattern analysis on MRI data, with equations for analysis and synthesis as well as description of how the TBM pipeline can be used for discrimination, regression, and unsupervised learning}
\item{Demonstration of TBM on real world neuroimage analysis problems showing the advantage of a generative technique in identification of morphologic changes as well as visualization compared to current morphometry techniques}
\end{itemize}

The remainder of this paper is organized as follows. In Section \ref{sec:review}, we summarize key theoretical results and present equations for forward and inverse TBM transformation. In Section \ref{sec:proposed}, we present our TBM solver suitable for MRI data. Section \ref{sec:models} describes a framework for regression, discrimination, and blind signal separation tasks in the transform domain. In Section \ref{sec:method}, experimental methodology is presented. Section \ref{sec:results} presents results showing robustness of the proposed solver for 3D data and the ability of TBM to accurately assess dependent brain morphology changes with age in Section \ref{sec:results}, whereas the traditional approaches based on diffeomorphic anatomic registration through exponentiated lie algebra algorithm (DARTEL) fails to detect these changes \cite{ashburner2007fast}. Finally, in Section \ref{sec:discussion}, we present discussion of the results and in Section \ref{sec:conclusion}, we conclude this paper. The Appendix material details the derivation and experimental validation for our solver in Section \ref{sec:proposed}. 

\section{Optimal Mass Transport for Signal Transformation}
\label{sec:review}

This section summarizes key theorems related to optimal mass transport, equations for signal transformation using TBM, and OMT minimization.  

\subsection{Overview of optimal transport theory}
Let $\Omega$ be a measurable space. Let $\mu$ and $\sigma$ be probability measures defined on $\Omega$, with corresponding positive probability densities $I_1$ and $I_0$, respectively. A mass preserving transform $f$ that pushes $\sigma$ to $\mu$, or $f_\#\sigma =\mu$, satisfies the following, 

\begin{eqnarray}
\int_Ad\sigma(x)=\int_{f(A)} d\mu(x),~~~\forall A\subset \Omega.
\label{eq:mp}
\end{eqnarray}

\begin{figure}[h!]
\centering
  \includegraphics[width = 0.6\columnwidth]{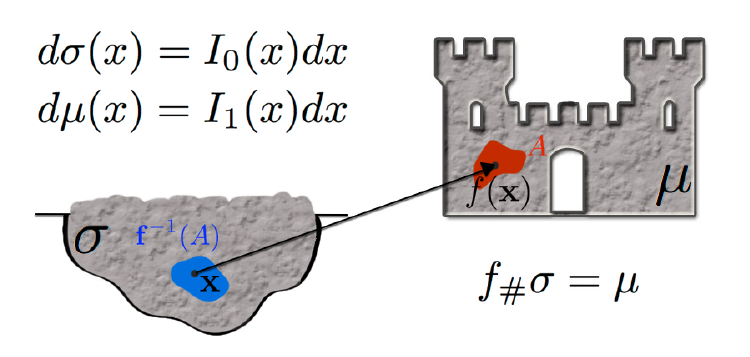}
    \caption{\footnotesize{The optimal coupling is sought between units of mass in the pile of dirt and units of mass in the castle such that the transportation cost is minimized}}
\label{fig:dirtpile}
\end{figure}

Figure \ref{fig:dirtpile} illustrates $\mu$ and $\sigma$, as well as the map $f_\#\sigma = \mu$. Such a mass preserving (MP) mapping $f$ is in general not unique; in fact, infinitely many MP mappings may exist that satisfy Equation \ref{eq:mp}. However, we are interested in finding the MP mapping that is optimal in the sense of \textit{mass transport}, which we will define further in Equation \ref{eq:monge}. Let $MP$ be the set of all such mass preserving mappings, $MP:= \{ f:\Omega\rightarrow\Omega | f_\#\sigma=\mu \}$. The optimal MP mapping in the mass transport sense can be written according to Monge's formulation, which minimizes the following cost function,

\begin{eqnarray}
\min_{f\in MP} \int_\Omega c(x,f(x)) d\sigma(x)
\label{eq:monge}
\end{eqnarray}

Here, $c:\Omega\times\Omega\rightarrow \R^+$ is the cost functional. The functional $c$ measures mass transportation cost and is often chosen to be the $L^p$-norm for which Equation \eqref{eq:monge} becomes the $L^p$-Wasserstein distance. The $L^2$-Wasserstein distance, $c(x,y)=|x-y|^2$, in particular has attracted rich attention in the image analysis, computer vision, and machine learning communities. For $c(x,y)=|x-y|^2$, Br\'{e}nier \cite{brenier1991} showed that there exists a unique optimal transportation map $f\in MP$ for which, 

\begin{eqnarray}
\int_\Omega  |x-f(x)|^2 d\sigma(x)\leq \int_\Omega |x-g(x)|^2 d\sigma(x),\nonumber\\ \forall g\in MP 
\end{eqnarray} when (i) $\Omega=\R^n$ and the probability measures have finite second-order moments (i.e. their densities vanish in the limit),
\begin{eqnarray*}
\int_\Omega |x|^2d\mu(x)<\infty~~\text{and}~~\int_\Omega |x|^2d\sigma(x)<\infty,
\end{eqnarray*} 
and (ii) when $\sigma$ is absolutely continuous with respect to Lebesgue measure. Note that for certain measures (e.g. when $\sigma$ is not absolutely continuous) the Monge formulation of the optimal transport problem is ill-posed in the sense that there is no transport map that rearranges $\sigma$ into $\mu$. In such scenarios Kantorovich formulation of the problem is preferred.

Moreover, Br\'{e}nier showed through polar factorization \cite{brenier1991} that the transport map $f$ must be the gradient of a convex function $\phi:\Omega\rightarrow\R$, $f=\nabla \phi$. The preceding property implies that when $\Omega$ is a convex and connected subset of $\R^n$, the optimal transport map is \textit{curl free}. 

\subsection{Linear optimal transport analysis framework}
\label{ssec:analysis_synthesis}

By considering magnetic resonance images to be smooth density functions, the similarity in \textit{spatial distribution} of two tissues can be quantified based on the $L^2$-Wasserstein distance. Any MRI modality that generates scalar intensity maps (i.e. T1-weighted, T2-weighted, FLAIR, fractional anisotropy, etc.) is amenable to analysis by the TBM framework. In this work, we analyze T1-weighted images, where treating images as densities enables comparison between images where the absolute intensity value may not be physically meaningful.

Consider a set of magnetic resonance images $I_1,...,I_K: \Omega \rightarrow \R^+$, corresponding to experimental subjects $1,...,K$, where $\Omega=[0,1]^3$, the images are first intensity normalized to produce densities such that 
\begin{equation}
\int_\Omega I_m(x)dx=1.
\end{equation}

where $m \in {1,...,K}$. A common reference image $I_0$, is chosen and the optimal transport mappings are calculated from the reference image to each subject's MRI, $I_m$. Let $f_m:\Omega\rightarrow \Omega$ be a mass preserving mapping from $I_0$ to $I_m$. Then, the analysis equation \cite{kolouri2016continuous} that transforms images to their corresponding representation in transform domain can be written based on 

\begin{equation}
\begin{split}
f^*_m(x) = arg \min_{f_m \in \MP} \int_\Omega (f_m(x)-x)^2 I_0(x)dx, \\
\text{s.t.~~~~~~} \det(Df_m(x)) I_m(f_m(x))=I_0(x) ~~~~~~~\text{for}~~\forall x \in \Omega
\label{eq:analysis}
\end{split}
\end{equation}

Here, $Df_m$ is the Jacobian of the mapping $f_m$ and $\MP$ is the family of all mass preserving mappings from $I_0$ to $I_1$. The existence of a unique solution $f^*_m$ to above optimization was shown by Br\'{e}nier \cite{brenier1991}. 

\begin{figure}
\includegraphics[width = \columnwidth]{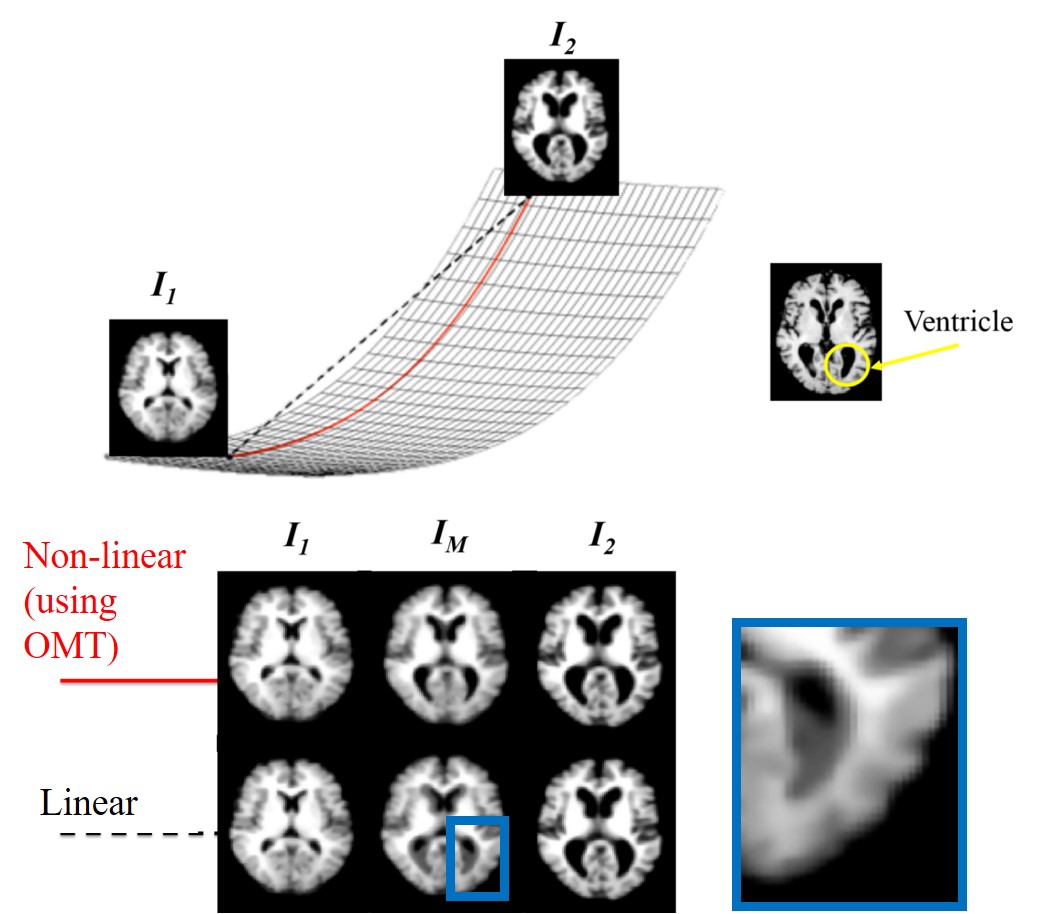}
\caption{Compared to the simple Euclidean distance (black), the OMT distance (red) between images defines a nonlinear distance metric between a pair of images, represented by the red arc on the manifold. Compared to simple Euclidean interpolation to estimate the middle image $I_M$, which leads to artifacts in the ventricles and other artifacts, the nonlinear OMT-based distance appears to better capture the natural structure of the brain.}
\label{fig:manifold}
\end{figure}

The transport maps $f^*_m(x)$ are vector fields that define the direction and amount of mass transport needed to morph $I_m(x)$ into $I_0(x)$. OMT defines a nonlinear distance metric, as Figure \ref{fig:manifold} shows, where the arcs, or geodesics, on the manifold between two images $I_0$ and $I_m$ correspond to nonlinear OMT distances and are represented by $f^*_m(x)$. The metric space defined by the OMT-based distance metric is a Riemannian manifold, which is equipped with an inner product. Thus, projecting the manifold locally at $I_0$ to the tangent space maps the geodesics $f^*_m$ to linearized versions in the tangent space, called the linearized optimal transport (LOT) metric. 

Then, it can be shown that $\hat{I}_m(x)=(f^*_m(x)-x)\sqrt{I_0(x)}$  provides a natural isometric linear embedding for image $I_m$ with respect to the LOT \cite{wang2013linear, kolouri2016continuous}. This linear embedding is generative, thus, any arbitrary point in the LOT space can be directly inverted and visualized in the image domain \cite{wang2013linear, kolouri2016continuous, kolouri2016radon} as a new image according to the synthesis equation \cite{wang2013linear, kolouri2016continuous} 

\begin{equation}
\begin{split}
I(x) = det(Df^{-1}(x))I_0(f^{-1}(x)) \\
\text{where~} f^{-1}(x) \text{~is the inverse mapping of~} f(x)
\end{split}
\label{eq:synthesis}
\end{equation}

LOT is a powerful technique for image analysis because simple Euclidean operations on the LOT-transformed embeddings, such as linear classification and linear regression correspond to nonlinear operations on the OMT manifold. Thus complex, spatially diffuse, nonlinear morphologic shifts in the image domain can be captured by simple Euclidean operations in the LOT domain, facilitating discovery of trends as well as visualization of these trends through inverse transformation \cite{kolouri2016continuous}. 

\section{Proposed approach}
\label{sec:proposed}

We have developed a method for solving optimal transport that enables the transport-based morphometry technique to be extended to large 3D volumetric images such as MRI. The authors offer this OMT approach as a viable option for carrying out TBM transformation for 3D volumetric images. While numerical OMT is a vast field, a detailed review or evaluation of OMT algorithms in general is beyond the scope of this paper. 

\subsection{Variational formulation of the problem}
\label{ssec:formulation}

In order to find the optimal transport map, we reformulate the minimization in \eqref{eq:monge} by relaxing the MP constraint. Assuming that $\Omega$ is a convex and connected subset of $\R^n$, as it is the case for most image analysis problems (i.e. $\Omega=[0,1]^n$), and assuming that the probability measures $\mu$ and $\sigma$ are atomless and absolutely continuous, we can write the differential counterpart of Equation \eqref{eq:mp} as, 
\begin{eqnarray}
det(Df(x)) I_1(f(x))=I_0(x) ,~~~\forall f\in MP
\end{eqnarray}
where $D$ is the Jacobian matrix, and $det(.)$ denotes the determinant operator.  The minimization in \eqref{eq:monge} for $c(x,y)=|x-y|^2$ can  first be relaxed into the following optimization problem, 
\begin{eqnarray}
&\operatorname{argmin}_f& \frac{1}{2}\int_\Omega |x-f(x)|^2 I_0(x)dx\nonumber\\
&s.t.& ~\|det(Df) I_1(f)-I_0\|^2\leq \epsilon
\label{eq:relaxed}
\end{eqnarray}
for some small $\epsilon>0$. Next, we use the result from Br\'{e}nier's theorem which states that the optimal transport map is a curl free mass preserving map. Therefore we propose to modify the optimization problem in Equation \eqref{eq:relaxed} by regularizing the objective function with the curl of the mapping, $f$,
\begin{eqnarray}
&\operatorname{argmin}_f&  \frac{1}{2}\int_\Omega |x-f(x)|^2 I_0(x)dx+\frac{\gamma}{2} \int_\Omega | \nabla\times f(x)|^2 dx\nonumber\\
&s.t.& \|det(Df) I_1(f)-I_0\|^2\leq \epsilon
\label{eq:relaxed2}
\end{eqnarray}
where $\gamma$ is the regularization coefficient and $\nabla \times (.)$ is the curl operator. We note that modifying \eqref{eq:relaxed} to penalize the objective with the curl does not change the optimal solution, but solving \eqref{eq:relaxed2} in practice helps guide the solution toward the curl free map. We can relax the optimization problem above further and write it as a regularized (or penalized) unconstrained optimization problem, 
\begin{eqnarray}
\operatorname{argmin}_f  &&\frac{1}{2}\int_\Omega |x-f(x)|^2 I_0(x)dx+\frac{\gamma}{2} \int_\Omega | \nabla\times f(x) |^2 dx\nonumber\\ &+&\frac{\lambda}{2} \int_\Omega (det(Df(x)) I_1(f(x))-I_0(x))^2 dx\nonumber\\
\label{eq:proposed}
\end{eqnarray}
Hence the formulation above contains terms explicitly signifying properties of MP mapping $\|det(Df)I_1(f)-I_0\|^2$ and a curl-free mapping $\|\nabla \times f\|^2$.  The last term implicitly penalizes mappings that are not diffeomorphic when $det(Df(x))$ crosses zero. 

The optimization problem in Equation \eqref{eq:proposed} is not a convex problem. We use a multiscale variational optimization technique to help guide the solution toward the global optimum. We will see in the results section that the multiscale scheme is able to achieve solutions comparable to that obtained using convex methods when they apply. Section \ref{ssec:multiscale} describes the multiscale variational solver we devise for the optimization in \eqref{eq:proposed}.

\subsection{Euler-Lagrange equations}
\label{ssec:grad}

The objective function in \eqref{eq:proposed} can be written as, 
\begin{eqnarray}
M(f)&=\int_\Omega \L(x,f(x),Df(x))dx.
\label{eq:standard}
\end{eqnarray}
The Euler-Lagrange equations for the transport field $f$ then are of the form, 
\begin{eqnarray}
\frac{d M}{d f^i}=\frac{\partial \L}{\partial f^i} - \sum_{k=1}^n \frac{d}{d x^k}(\frac{\partial \L}{\partial f_{x^k}^i}),~~~ i=1,...,n
\end{eqnarray}
where the superscripts denote the coordinate index for the vectors, and the subscripts denote partial derivatives, $f_{x^k}^i= \frac{\partial f^i}{\partial x^k}$. Writing the Euler-Lagrange equations for the objective function in \eqref{eq:proposed} leads to, 
\begin{eqnarray}
\frac{d M}{d f}&=&(f-id)I_0+\lambda(det(Df)\nabla I_1(f) 
\nonumber\\ &-& \nabla \cdot (adj(Df)I_1(f)))I_{error}\nonumber\\ 
&+& \gamma (\nabla \times \nabla \times f)
\label{eq:EL}
\end{eqnarray}
where $id(x)=x$ is the identity function, $adj(.)$ denotes the adjugate operator, $\nabla\cdot (.)$ is the divergence operator, and $I_{error}=det(Df)I_1(f)-I_0$. The derivation for the equation above is presented in the Appendix. Equation \eqref{eq:EL} is a key result from our formulation. The complexity for computing each gradient descent update step here is $\O(NlogN)$, where N is the number of pixels or voxels in the image. The computational complexity is determined by the cost of computing gradients $\O(N)$ in \eqref{eq:EL}, but is dominated by the cost of cubic interpolation in computing the $det(Df)I_1(f)$ term. 

\subsection{Multiscale accelerated gradient descent}
\label{ssec:multiscale}

\begin{figure}[t]
  \includegraphics[width=\columnwidth]{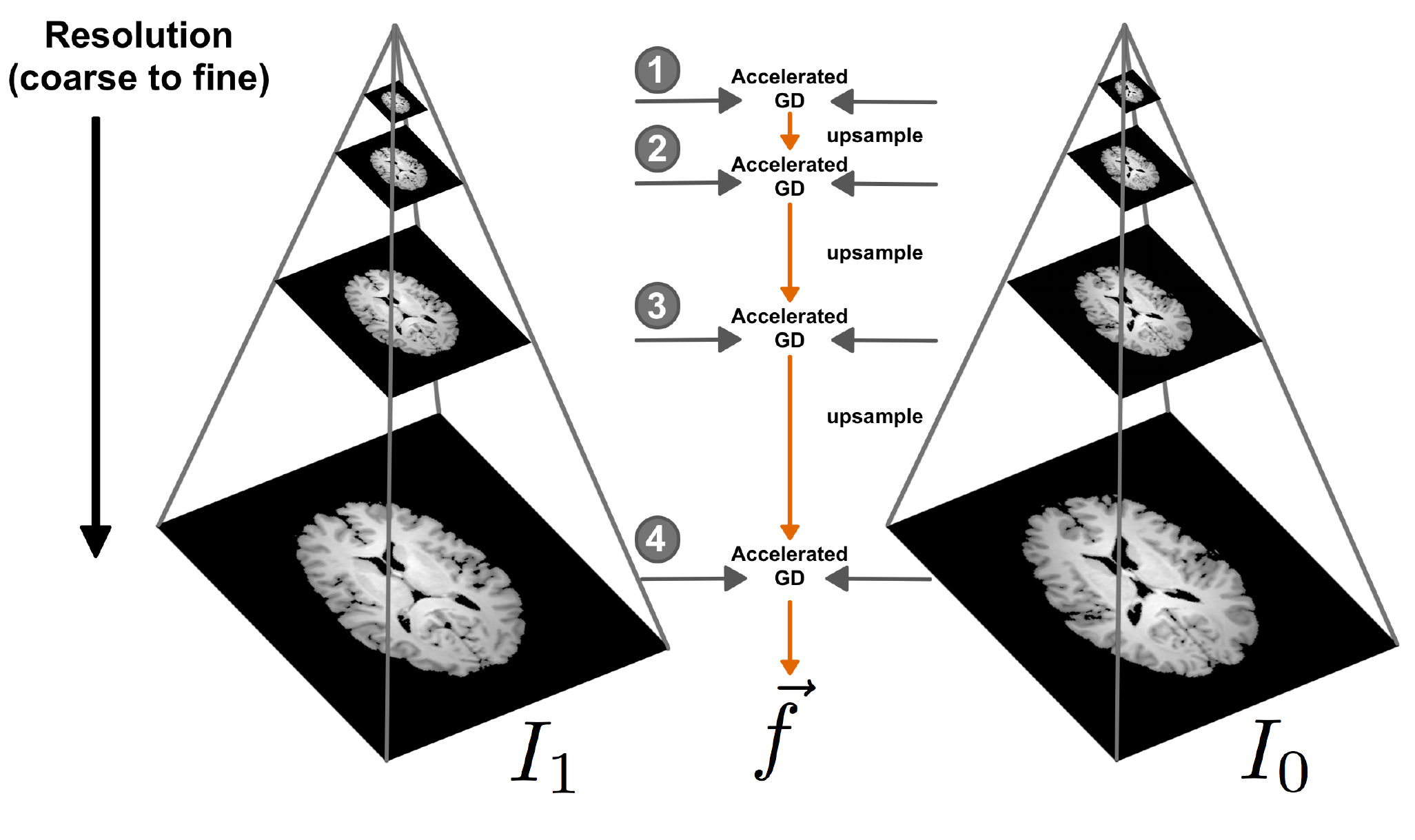}
  \vspace{-2em}
   \caption{\footnotesize{The schematic of the multiscale approach devised in this paper. The solution to the accelerated gradient descent is first calculated at a coarse level and then refined as the optimization proceeds.}}
   \label{fig:multiscale}
\end{figure}

We can guide the solution toward the globally optimal solution by a multiscale scheme as depicted in Figure \ref{fig:multiscale}. Nesterov's accelerated gradient descent method \cite{nesterov2007gradient} is used at each scale to find the corresponding optimal transport map from Equation \eqref{eq:proposed}. The optimal transport map is then interpolated and used as the initial point for the accelerated gradient descent method in the next scale (finer scale). 

The accelerated gradient descent update for $k$'th iteration ($k>1$) at each scale is as follows, 
\begin{eqnarray}
\left\{ \begin{array}{l}
g^k=f^{(k-1)}+\frac{k-2}{k+1}(f^{(k-1)}-f^{(k-2)})\\
f^{k}=g^k-\alpha_k \frac{d M(g^{(k)})}{d f}
\end{array}\right.
\end{eqnarray}
where $\alpha_k$ is the gradient descent step size, and is automatically chosen at each gradient descent update such that the maximum displacement is fixed. The update at k = 1 is the usual gradient descent update. 

Here, we have presented a viable approach for computing optimal transport minimization for MRI datasets, enabling a transport-based morphometry approach with radiology images. 

\section{Modeling shape and appearance of the brain}
\label{sec:models}

As previous work has established that treating signals as densities increases their separability in the transform domain \cite{wang2013linear, wang2011optimal, basu2014detecting, park2017cumulative, kolouri2016continuous, kolouri2016radon}, we describe a framework for regression, discrimination, and blind signal separation in the transport space. 

\begin{figure}
\includegraphics[width = \columnwidth]{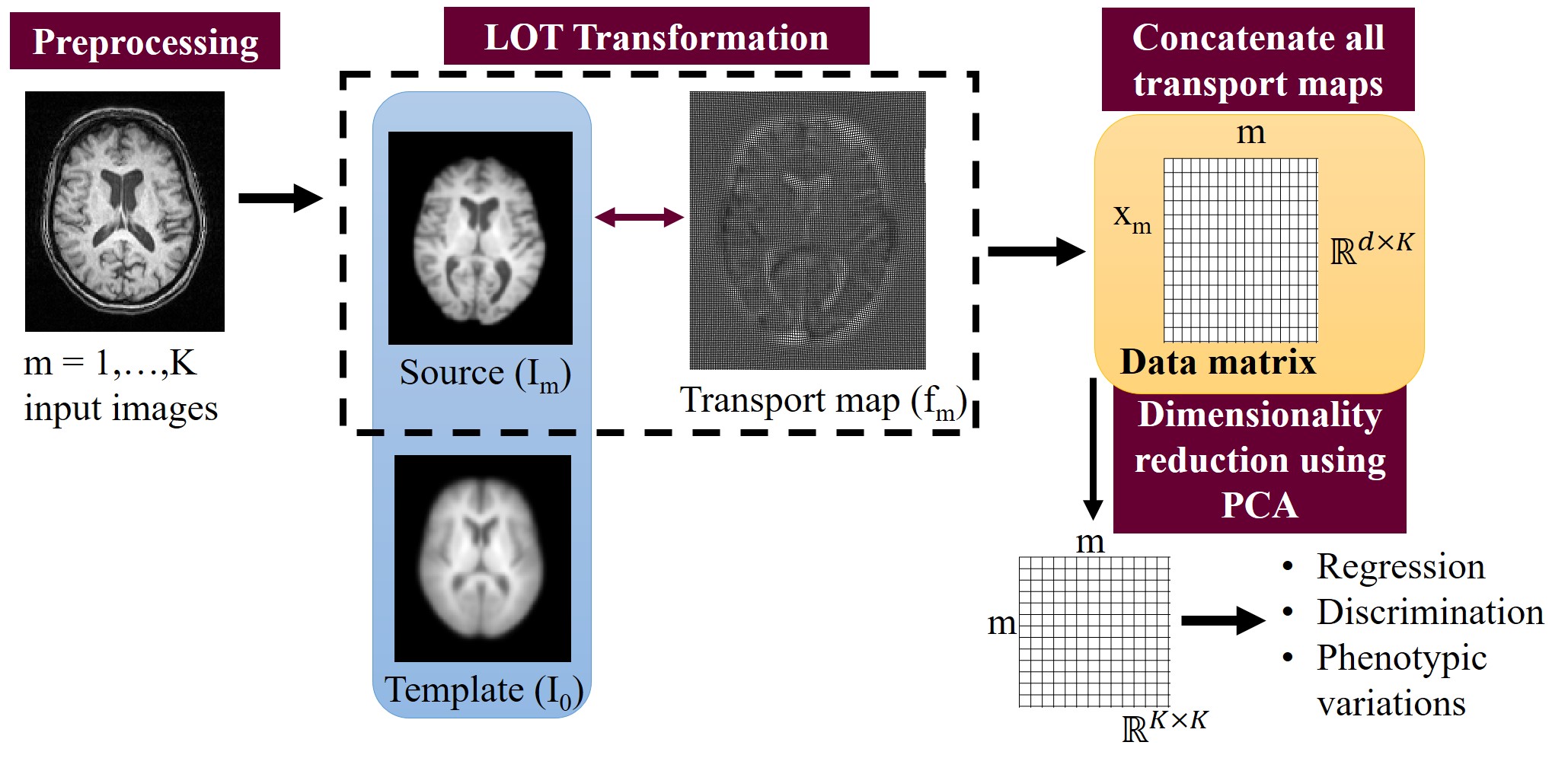}
\caption{System diagram. Images are first skull-stripped, intensity normalized and affine registered. A series of transport maps is computed using LOT transformation, and subsequent pattern analysis is performed in the transport space. Inverse LOT transformation provides visualization of the regression, discrimination, or principal component directions in the transport space for physical interpretation.}
\label{fig:diagram}
\end{figure}

The data matrix $X \in \mathbb{R}^{d \times K}$ stores the vectorized transport maps $x_m$ corresponding to each subject $m \in 1,...,K$ where $d$ is the number of elements in the vectorized transport map and $K$ is the number of subjects. Figure \ref{fig:diagram} is the system diagram that illustrates the LOT transformation pipeline. In practice, the analysis is performed on the dimensionality-reduced data matrix for ease of computation.

\subsection{Regression and correlation analysis with a clinical variable}
\label{ssec:regression}
The influence of an independent clinical variable $v \in \mathbb{R}^{K \times 1}$ on brain tissue distribution can be investigated by computing the direction in the transport domain $w_{corr}$ such that the linear correlation with age is maximized according to \eqref{eq:regression} \cite{basu2014detecting}. Here, $X$ represents the reduced-dimension data matrix. 

\begin{equation}
w_{corr} = arg \max_w \frac{w^T X v}{\sqrt{w^T w}} = \frac{X v}{\sqrt{v^T X^T X v}}
\label{eq:regression}
\end{equation} 
Here, the direction $w = \bar{x} + \nu w_{corr}$ is a vector field that represents the direction and magnitude by which tissue is re-distributed due to $v$ and $\nu$ represents the increment or decrement to sample along the maximally correlated direction. Pearson's correlation coefficient is computed on centered $v$ and $X$. 

The images corresponding to the computed direction $w$ can be visualized through inverse TBM transformation by Equation \eqref{eq:synthesis} and illustrate the morphology that is associated with outcome $v$.

\subsection{Discriminant analysis to differentiate groups of subjects}
\label{ssec:discriminant}
Another class of problems facilitated by the TBM technique is that of discriminating classes based on MRI appearance, such as the one posed in Figure \ref{fig:unfit}. For these problems, penalized linear discriminant analysis (PLDA) \cite{wang2011penalized} performed in the transport domain can find the direction in transport space that maximally separates $C$ classes. The PLDA direction is given by \eqref{eq:PLDA}

\begin{equation}
w_{PLDA} = arg \max_{||w||=1} \frac{w^TS_Tw}{w^T(S_W + \alpha I)w}
\label{eq:PLDA}
\end{equation} 

where $S_T = \frac{1}{M}\sum_m(x_m-\bar{x})(x_m-\bar{x})^T$. Here, $\bar{x} = \frac{1}{M} \sum_{m=1}^M x_m$.

The within-class scatter matrix is $S_W = \sum_C \sum_{n \in C} (x_n-\bar{x_c})(x_n-\bar{x_c})^T$. The parameter $\alpha$ controls the tradeoff between the traditional linear discriminant analysis (LDA) direction and one that lies in the principal component analysis (PCA) subspace. The parameter $\alpha$ can be chosen by plotting the stability of the subspace as a function of $\alpha$. 

Sampling along and inverting the direction $w_{PLDA}$ yields images showing the typical morphology of a class and how it changes as one progresses from one class to another.

\subsection{Visualizing principal phenotypic variations in the brain}
\label{ssec:PCA}
Given the covariance matrix $S_T$ defined in Section \ref{ssec:discriminant}, the principal components are given by the eigenvectors of $S_T$. The eigenvectors represent the directions in the transport space that capture the main modes of variability in the dataset \cite{basu2014detecting}.

The factorization in Equation \eqref{eq:PCA} gives both the principal components and eigenvalues, where the diagonal components of $\Sigma$ represent the variance for each principal component. 

\begin{equation}
S_T = U \Sigma U^T
\label{eq:PCA}
\end{equation}

For high dimensional data, the covariance matrix can be implicitly represented using the approach in \cite{cootes1995active}. Each principal component can be inverted and visualized to yield the principal phenotypic variations that comprise the images in the dataset. 

\section{Computational experiments}
\label{sec:method}

Here we describe image acquisition, preprocessing, morphometry analysis, and statistical learning steps. The code was prototyped in MATLAB (MathWorks, Natick, MA) using built-in libraries. 

\subsection{Datasets}

\subsubsection{MRI pattern analysis using transport-based morphometry}

The ability of transport-based morphometry to aid in regression, discrimination, and signal separation tasks was assessed on images of 135 healthy subjects, ranging in age from 58 to 81 years (mean age 66.6 years, standard deviation 5.9 years). Both male and female subjects are included. T1-weighted brain images were collected using a 3D Magnetization Prepared Rapid Gradient Echo Imaging (MPRAGE) protocol with 144 contiguous slices. Images were acquired on a 3 T Siemens Allegra scanner with repetition time = 1,800 ms, echo time = 3.87 ms, field of view (FOV) 256 mm, and acquisition matrix $192 \times 192$ mm , flip angle = 8 \cite{erickson2009aerobic}. These images provide an expanded dataset of older subjects on which age-related brain morphology can be investigated.

\subsection{Multiscale variational optimal transport}

\subsubsection{Image preprocessing}
\label{sssec:preprocessing}
Images were skull-stripped and affinely registered to the MNI template using Statistical Parametric Mapping (SPM) software version 12 \cite{spm12}. Images were normalized so that the sum of intensities was equal in both images (equal \textit{mass}). By normalizing to a large positive number, $10^6$, numerical precision errors resulting from computations with small numbers are avoided. We also add a small constant 0.1 to the normalized images and renormalize so that they are strictly positive \cite{chartrand2009gradient} to ensure that the OMT problem is well-posed. 

\subsection{Experiment 1: Modeling the effects of aging on brain tissue distribution} 

Regression analysis was performed to assess the relationship between brain tissue distribution and age using the approach outlined Section \ref{ssec:regression}. The common reference image $I_0$ was computed by the Euclidean average of all the subjects. Statistical significance of the computed direction is assessed using permutation testing with T = 1000 tests. 

The reults of regression analysis in the transport space were compared to those obtained using deformation-based analysis. The DARTEL \cite{ashburner2007fast} toolbox in SPM12 \cite{spm12} was used to compute deformation fields. DARTEL is commonly used to perform standard VBM and DBM analysis. Images were skull-stripped, segmented, and affine registered to the MNI template similar to the OMT procedure before non-rigid registration by performed by DARTEL. The most correlated direction was computed on the tissue density maps of DARTEL-registered images using Equation \eqref{eq:regression} for VBM and using the deformation fields for DBM analysis. Modulated versions are used to compensate for the effects of spatial normalization \cite{mechelli2005voxel}. 

The TBM analysis is also performed on segmented gray matter and white matter tissue maps separately to enable comparison to VBM.

\subsection{Experiment 2: Assessing the effects of aerobic fitness on brain health}

Discriminant analysis between high aerobic fitness vs. low aerobic fitness groups is performed using the PLDA approach in transport space described in Section \ref{ssec:discriminant}. Aerobic fitness is measured by $vO_2$ L/min. The individuals were grouped into low-fit and high-fit groups based on those who had a $vO_2$ L/min greater than one standard deviation above the mean (high-fit: n = 22) and lower than one standard deviation below the mean (low-fit: n = 16).

Discriminant analysis using TBM was compared with that performed using deformation fields instead of transport maps generated by DBM. In VBM analysis, voxelwise comparison on modulated tissue maps was performed to seek the voxel clusters that are significantly different between the two classes using two-sample t-test, uncorrected for multiple comparisons or cluster thresholding.  

\subsection{Experiment 3: Visualizing principal phenotypic variations}

Unsupervised learning using PCA was performed to visualize the top three principal phenotypic variations in the transport space using the approach described in Section \ref{ssec:PCA}.

\section{Results}
\label{sec:results}

\subsection{Modeling normal variability in the brain}

\begin{figure}
\centering
\includegraphics[width=0.7\linewidth]{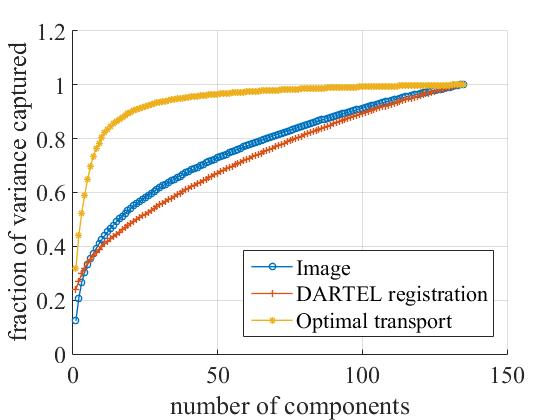}
\caption{\footnotesize{Compared to the models utilizing pixel-wise comparison (Eulerian and DARTEL registration), the model based on OMT is able to capture more of the variability in the dataset with fewer principal components.}} 
\label{fig:scree}
\end{figure}

Figure \ref{fig:scree} shows the fraction of variance captured by principal components of the image domain (raw voxel values after affine registration) compared to modulated DARTEL registration and optimal mass transport solved using the described approach. Fewer components are needed to represent more of the variance in the transport space than either for image domain or DARTEL pixelwise comparison. Therefore, the information about variability in the dataset appears to be better captured by examining tissue distribution using OMT rather than comparing tissue intensities individually, before or after nonrigid registration. The intuition for how OMT better capture variability in the dataset was previously illustrated by Figure \ref{fig:manifold}.

\subsection{Modeling the effects of aging on brain tissue distribution through TBM regression}

Aging is clinically known to be associated with tissue atrophy and disproportionate loss of tissue from frontotemporal regions \cite{brody1970structural}. In this section, TBM is compared with DBM and VBM in the ability to independently discover and model these changes.

\subsubsection{Assessing global changes}

\begin{figure}
\centering
\subfloat[]{\includegraphics[width=0.7\linewidth]{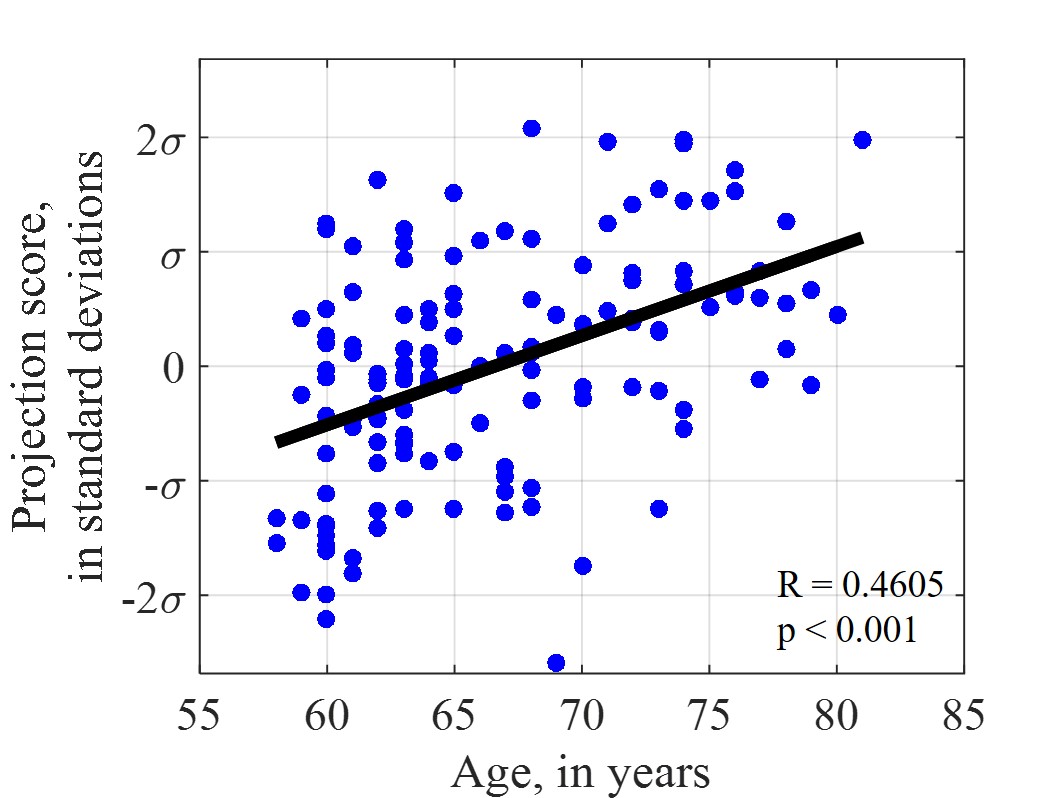}}
\newline
\subfloat[]{\includegraphics[width=0.8\linewidth]{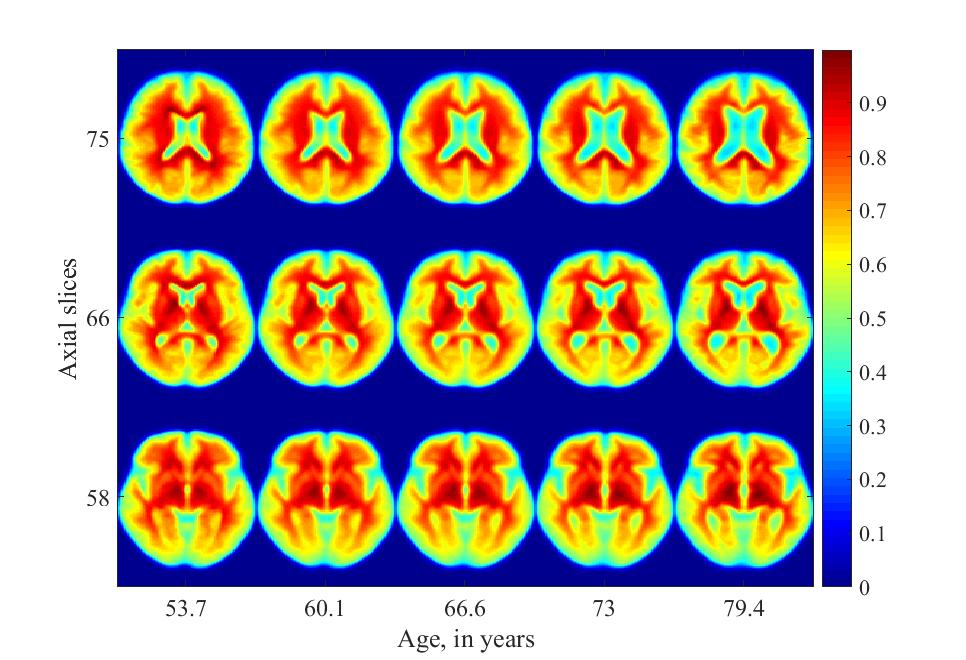}}
\caption{\footnotesize{(a) Projection of data onto the direction that maximizes linear correlation with age, (b) Visualization of changes in tissue distribution that are statistically dependent on age. The vertical axis shows various axial slices from a 3D dataset from rostral (towards head) to caudal (towards toe). The horizontal axis shows the effect of increasing age from left to right on that axial slice. We see enlarging ventricles, and global atrophy of both gray matter and white matter with increasing age.}}
\label{fig:HALT}
\end{figure}

The direction maximally correlated with age computed in the transport space using TBM is found statistically significant (Pearson's r = 0.4605, p $<$ 0.001). Figure \ref{fig:HALT}a shows the data when it is projected onto the maximally correlated direction, with each datapoint representing a subject's image. 

The most correlated direction shown in Figure \ref{fig:HALT}a can be inverted to visualize the dynamic changes in morphology underlying the aging process. Figure \ref{fig:HALT}b shows images generated by TBM inverse transformation (images are colorized to aid visual interpretation). We see that the changes captured by the TBM regression framework are well-corroborated by known changes in the clinical literature \cite{brody1970structural}. Specifically, the changes shown here are enlarging ventricles, especially in slices 75 and 66. There is global tissue thinning, and enlargement of the occipital horns of the lateral ventricles in slice 58. Normal anatomic landmarks characteristic of the brain are also clearly visible in Figure \ref{fig:HALT}b, such as the internal capsule in slice 66 and thalamus in slice 58. 

\begin{figure}
\centering
\subfloat[]{\includegraphics[width=0.7\linewidth]{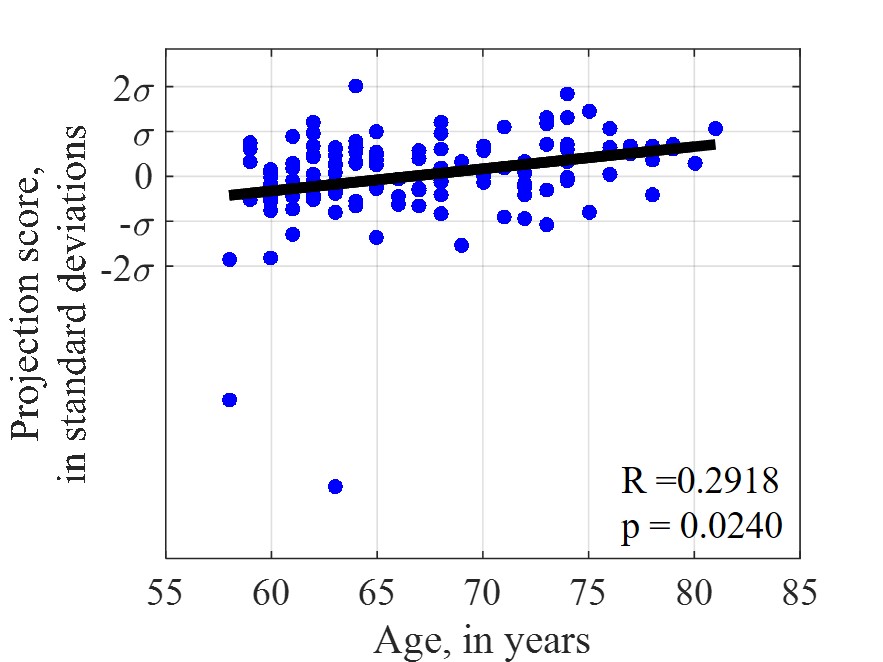}}
\newline
\centering
\subfloat[]{\includegraphics[width=0.8\columnwidth]{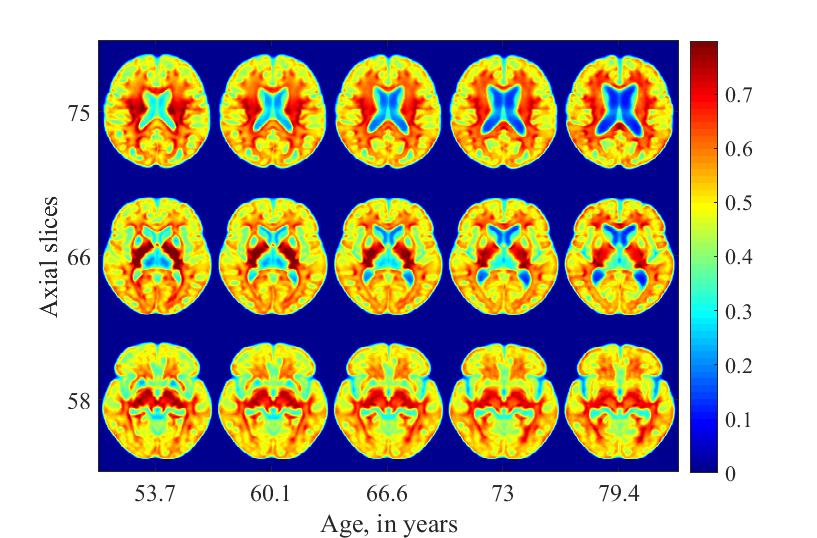}}
\caption{\footnotesize{Visualization of aging-related changes captured by DARTEL deformation fields used in deformation-based morphometry. Normal tissue texture is not well-modeled using a deformation-based approach.}}
\label{fig:DBM}
\end{figure}

We also compare the results of our TBM regression analysis with that using deformation fields generated by DARTEL, commonly used in DBM analysis.  The relationship between the deformation fields and age is found to be statistically significant (Pearson's r = 0.2918, p = 0.0240), as Figure \ref{fig:DBM}a shows, suggesting that there are significant shape changes with age that are captured by a DBM approach. Compared to the visualizations generated by TBM, we see that the those yielded by the deformation-based approach depict global shape changes but not texture changes. In all slices generated by deformation-based analysis, normal tissue landmarks are distorted. Especially, at the gray-white interface, there are is a ring-like texture that does not represent normal brain anatomy. Examining the associated images generated using the deformation-based approach, we see that while volume expansion of the ventricles is correctly identified, expected changes in tissue distribution are not well-captured using deformations alone. For example, in slice 75, the expected tissue thinning is not well-depicted in the frontal areas. In slice 66, there is an area of bilateral focal hyperintensity near the ventricles. This represents the distorted internal capsule that is correctly represented in slice 66 of Figure \ref{fig:HALT}b. Other normal landmarks such as thalamus and putamen are notably absent in slice 58 as well as the occipital horns of the lateral ventricles. Thus, while DBM can capture global shape changes, such as enlargement of ventricles, texture information is not well modeled and many normal landmarks and are distorted.

\subsubsection{Assessing gray matter and white matter changes}

\begin{figure}
\centering
\subfloat[]{\includegraphics[width=0.7\linewidth]{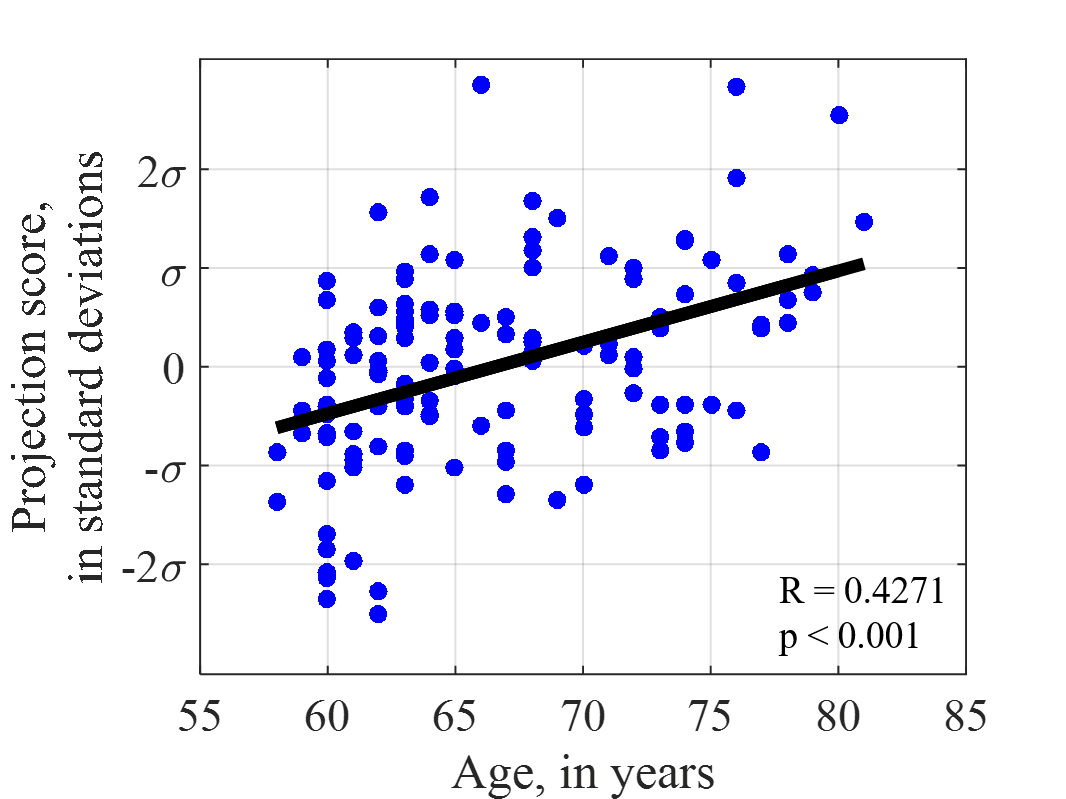}}
\newline
\centering
\subfloat[]{\includegraphics[width=0.8\linewidth]{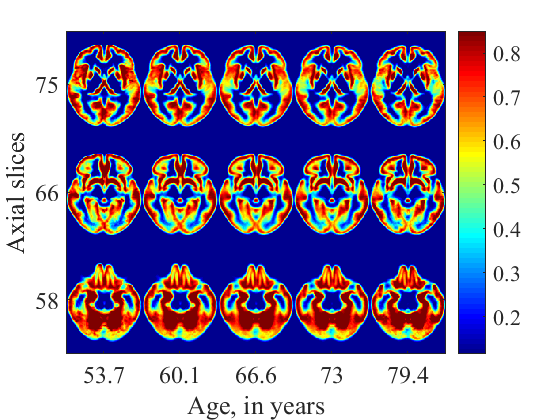}}
\caption{\footnotesize{Visualization of aging-related changes found by transport-based morphometry on gray matter channels. There is atrophy and loss of GM from temporal lobes.}}
\label{fig:TBM_gm_channels}
\end{figure}

Transport-based morphometry was also applied to explore the effect of age on gray matter and white matter maps separately. Figure \ref{fig:TBM_gm_channels}b shows the effect of age on gray matter distribution. The relationship is statistically significant with Pearson's r = 0.4271 and p$<$0.001. There is thinning of the gray matter tissue when progressing from a 53 year old brain to a 79 year old brain, most markedly in the temporal lobe as can be seen in slice 75. Atrophy can be seen in all the slices by enlargement of the spaces.

\begin{figure}
\centering
\subfloat[]{\includegraphics[width=0.7\linewidth]{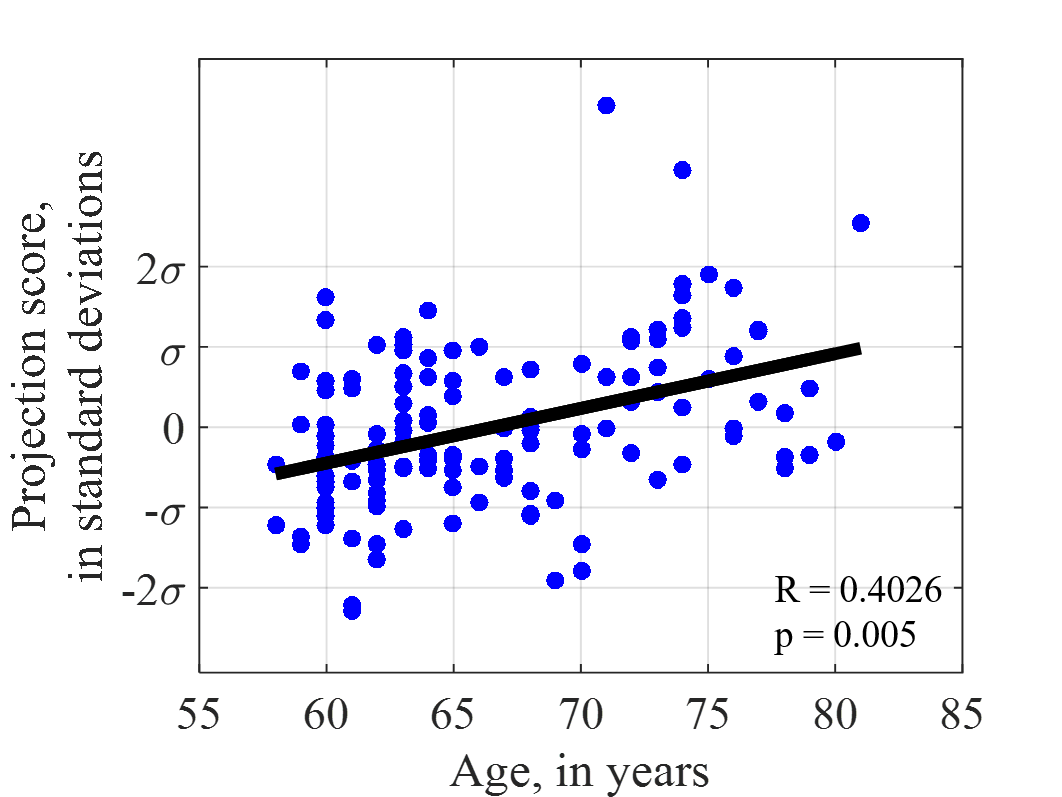}}
\newline
\centering
\subfloat[]{\includegraphics[width=0.8\linewidth]{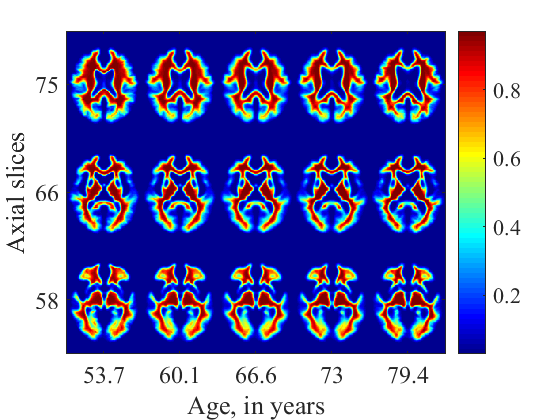}}
\caption{\footnotesize{Visualization of aging-related changes found by transport-based morphometry on white matter channels.TBM depicts overall atrophy and loss of white matter disproportionately from the frontal lobes.}}
\label{fig:TBM_wm_channels}
\end{figure}

The relationship between white matter distribution and age similarly shows atrophy and enlargement of the ventricles in Figure \ref{fig:TBM_wm_channels}. The relationship is statistically significant with Pearson's r = 0.4026 and p = 0.005. In addition, there appears to be disproportionate loss of white matter tissue from the frontal and temporal regions, which is best illustrated in slices 75 and 66 in Figure \ref{fig:TBM_wm_channels}b.


In contrast, regression analysis performed on the modulated density maps registered by DARTEL, used for VBM, was unable to find a significant relationship between age and either gray matter morphology (Pearson's r = 0.6787, p = 0.1980) or white matter morphology (Pearson's r = 0.4965, p = 0.3870).

\begin{figure}
\centering
\subfloat[]{\includegraphics[width=0.7\linewidth]{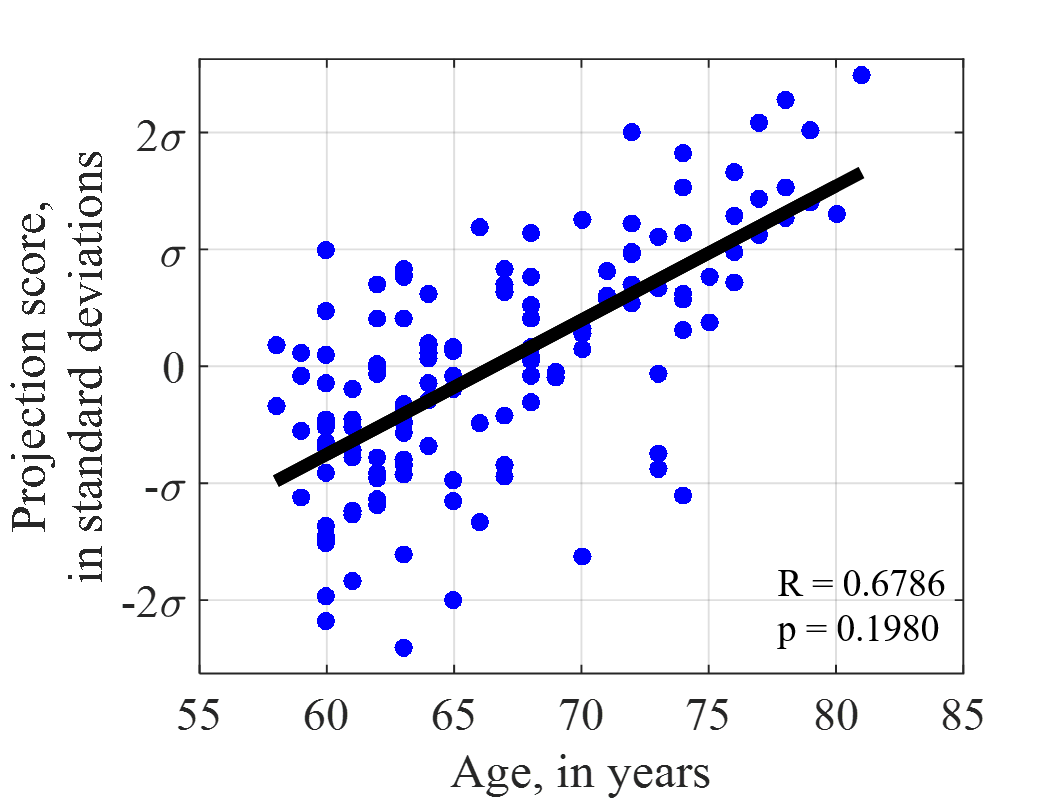}}
\newline
\centering
\subfloat[]{\includegraphics[width=0.8\linewidth]{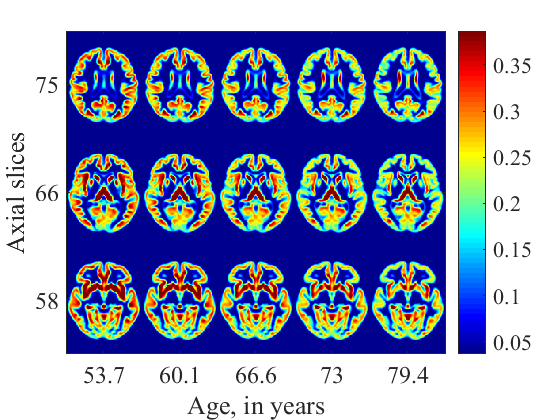}}
\caption{\footnotesize{Visualization of aging-related changes found by voxel-based morphometry on gray matter maps.}}
\label{fig:VBM_gm_channels}
\end{figure}

\begin{figure}
\centering
\subfloat[]{\includegraphics[width=0.7\linewidth]{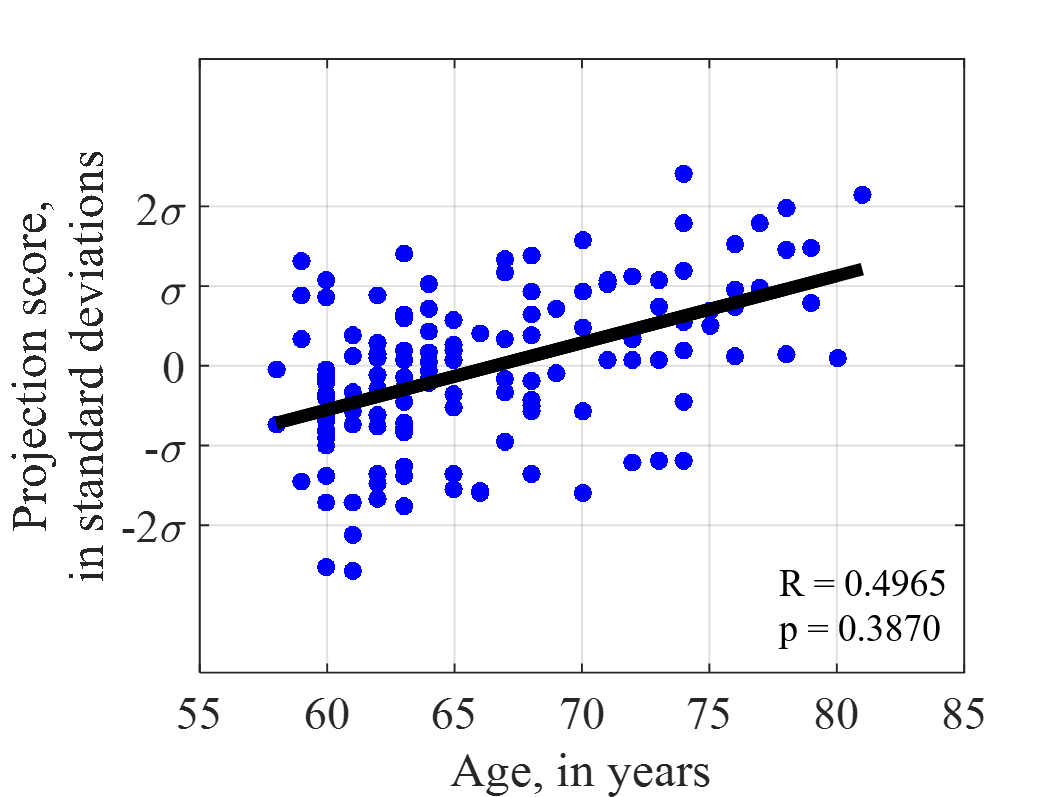}}
\newline
\centering
\subfloat[]{\includegraphics[width=0.8\linewidth]{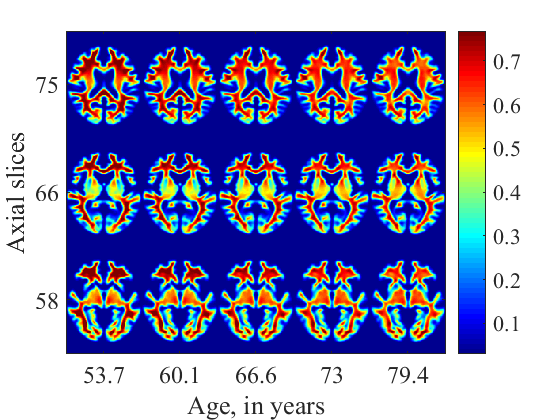}}
\caption{\footnotesize{Visualization of aging-related changes found by voxel-based morphometry on white matter maps.}}
\label{fig:VBM_wm_channels}
\end{figure}

Figures \ref{fig:VBM_gm_channels} and \ref{fig:VBM_wm_channels} show the images generated by attempting to fit a regression model on individual pixel values on a fixed grid. We see that in both cases, progressing from age 53 to age 79, the intensity at voxels in the cortical gray matter is shown to decrease. However, no gross differences in shape are depicted. Similarly, examining the white matter images generated by regression on VBM maps (Figure \ref{fig:VBM_wm_channels}), the intensity in the frontal white matter appears to grossly decrease, especially in slices 75 and 58. However, neither of these relationships were statistically significant, nor do they adequately depict atrophy and loss of tissue from frontotemporal regions. 

Overall, the known effects of aging on the brain are best assessed and depicted by the transport-based morphometry technique. A DBM approach does not adequately model tissue texture and VBM can identify intensity changes at fixed voxel locations, but these do not appear to be statistically significant when it comes to modeling the effect of aging on the brain tissue.

\subsection{Assessing the effects of aerobic fitness on brain health through TBM discrimination}

The effects of aerobic fitness on the brain are assessed by separating high-fit individuals from low-fit individuals using the PLDA approach for discriminant analysis, comparing the ability of TBM, DBM, and VBM to discover and visualize the interface between the two groups.

\begin{figure}
\centering
\subfloat[]{\includegraphics[width = 0.7\linewidth]{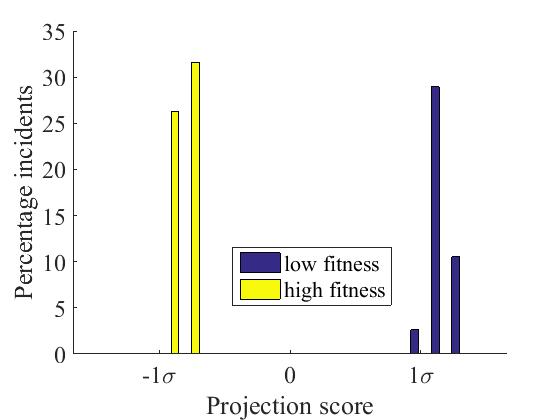}}
\newline
\subfloat[]{\includegraphics[width = 0.8\linewidth]{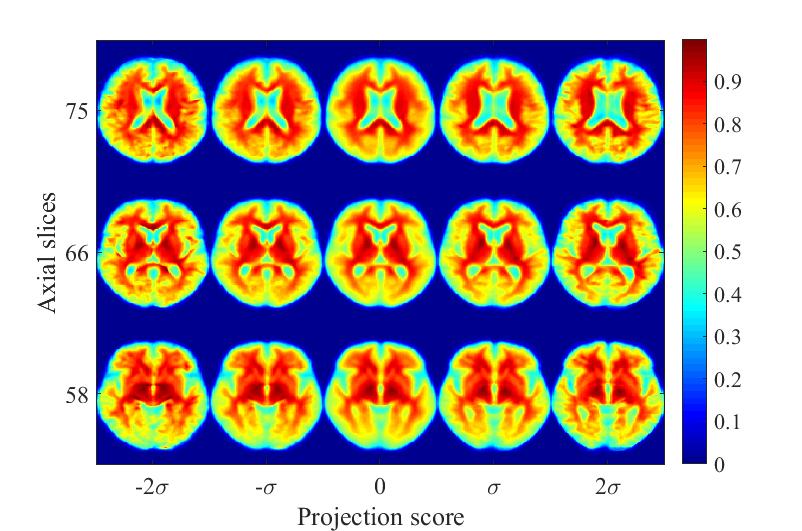}}
\caption{\footnotesize{(a) High-fit and low-fit individuals can be perfectly separated based on their transport maps given by TBM when projected onto the most discriminant direction computed by PLDA, (b) images illustrating the differences between high-fit and low-fit individuals generated based on transport maps.}}
\label{fig:histogram}
\end{figure}

\begin{figure}
\centering
\subfloat[]{\includegraphics[width = 0.7\linewidth]{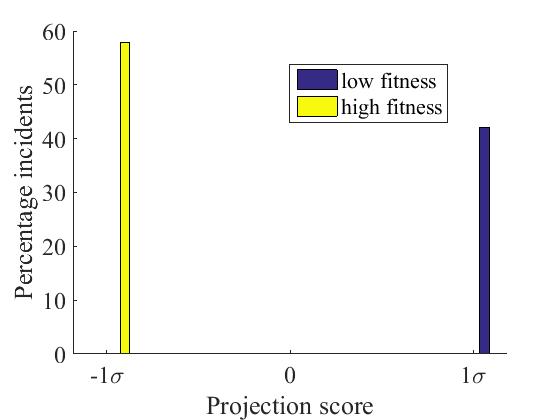}}
\newline
\subfloat[]{\includegraphics[width = 0.8\linewidth]{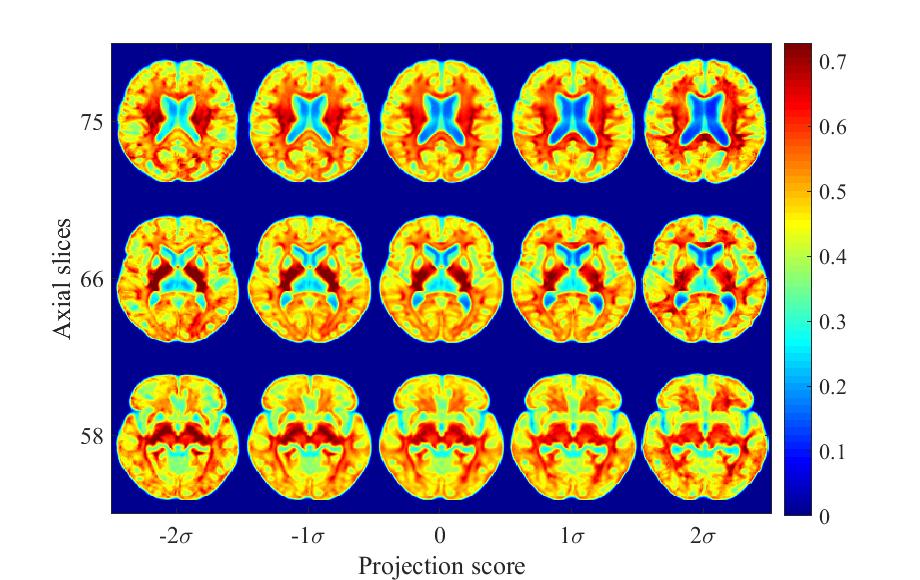}}
\caption{\footnotesize{(a) High-fit and low-fit individuals can be perfectly separated based on the deformation fields given by DARTEL when projected onto the most discriminant direction computed by PLDA, (b) modulated images illustrating the differences between high-fit and low-fit individuals generated based on deformation fields.}}
\label{fig:DBM_histogram}
\end{figure}

Clear separation in the training subspace is an expected result whether raw pixels, deformation fields or transport maps are used, as Figures \ref{fig:histogram}a and \ref{fig:DBM_histogram}a show, but when visualizing the interface between classes, TBM demonstrates clear advantages in physical interpretability. Visualizing the interface between the high-fitness and low-fitness groups using TBM in Figure \ref{fig:histogram}b, we see that brains corresponding to low-fit individuals appear to demonstrate changes in tissue distribution that are similar to those due to advancing age in Figure \ref{fig:HALT}b. Similarly, those individuals belonging to the high-fit group have brain morphology that appears to be resemble those of younger subjects in an older adult population as the ventricles appear smaller and tissue architecture in the frontotemporal regions are better preserved. Thus, it appears that fitness preserves areas of the brain that are affected in normal aging. 

Comparing the results to that obtained utilizing the deformation fields generated by DARTEL that are used in DBM analysis, the DBM visualizations show distortion of tissue topology. Figure \ref{fig:DBM_histogram}b appears to depict enlargement of ventricles with low fitness, but normal anatomic landmarks are not easily visualized, including the interface between gray and white matter. Additionally, texture variations are not well-assessed. 

\begin{figure}
\centering
\subfloat[]{\includegraphics[width = 0.7\linewidth]{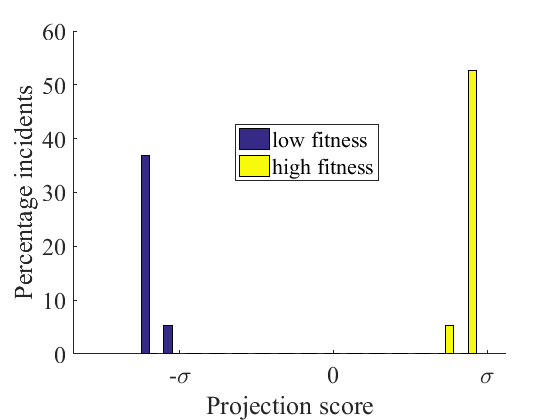}}
\newline
\subfloat[]{\includegraphics[width = 0.8\linewidth]{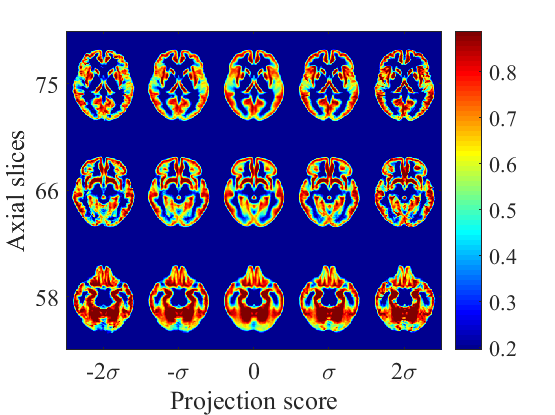}}
\caption{\footnotesize{(a) High-fit and low-fit individuals can be perfectly separated based on transport maps of gray matter when projected onto the most discriminant direction computed by PLDA, (b) modulated images generated by TBM depicting the gray matter differences between high-fit and low-fit individuals showing loss of tissue from temporal lobe in slice 75 with low fitness.}}
\label{fig:TBM_gray_PLDA}
\end{figure}

\begin{figure}
\centering
\subfloat[]{\includegraphics[width = 0.7\linewidth]{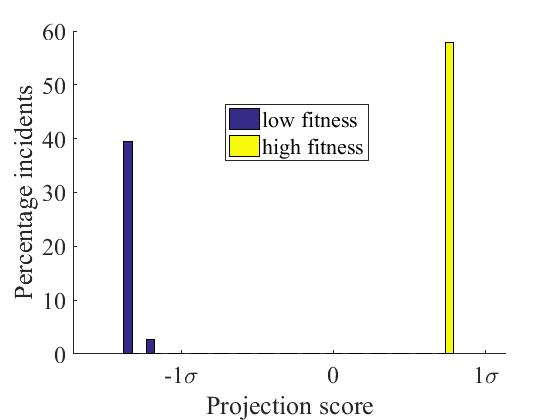}}
\newline
\subfloat[]{\includegraphics[width = 0.8\linewidth]{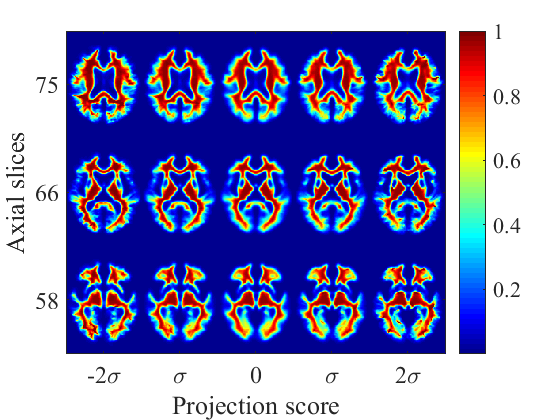}}
\caption{\footnotesize{(a) High-fit and low-fit individuals can be perfectly separated based on transport maps of white matter when projected onto the most discriminant direction computed by PLDA, (b) modulated images generated by TBM depicting the white matter differences between high-fit and low-fit individuals show that fitness appear to protect frontotemporal white matter architecture.}}
\label{fig:TBM_white_PLDA}
\end{figure}

The analysis is performed on gray matter and white matter maps separately as well in order to compare the performance of TBM with that of VBM. Figure \ref{fig:TBM_white_PLDA} and \ref{fig:TBM_gray_PLDA} show the results when transport-based morphometry is performed on white matter maps and gray matter maps individually.

The interface between the groups is visualized using TBM, which shows loss of temporal lobe gray matter with low fitness in Figure \ref{fig:TBM_gray_PLDA}b. White matter changes visualized by TBM shows loss of frontotemporal white matter with low fitness and enlarging ventricles in Figure \ref{fig:TBM_white_PLDA}b. The pattern of changes in brain tissue distribution seen is similar to that seen in aging.

\begin{wrapfigure}[23]{R}{0.55\columnwidth}\includegraphics[width=0.55\columnwidth]{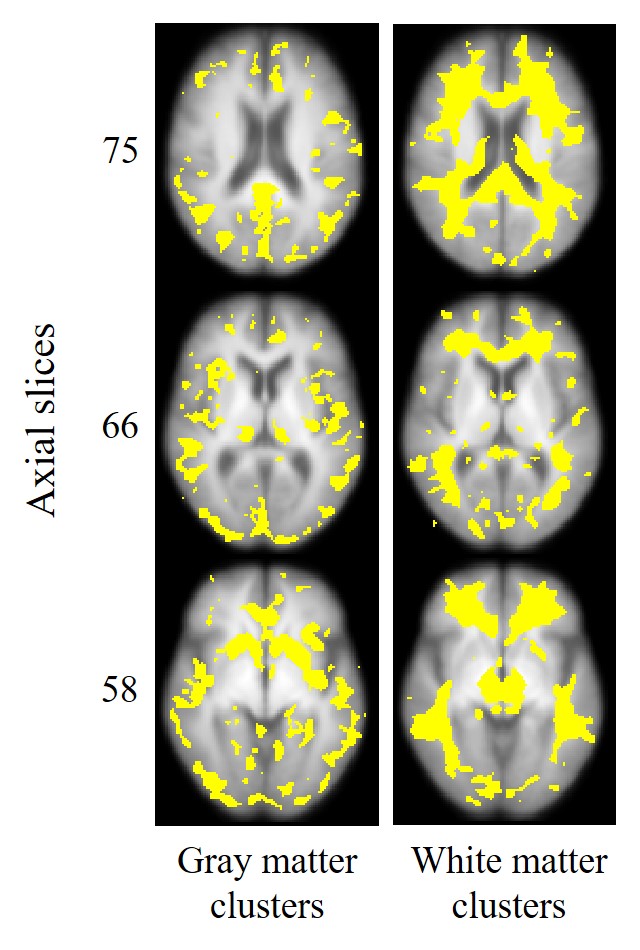}
\caption{\footnotesize{Heat maps showing the voxels on modulated gray matter and white matter densities whose intensity levels are significantly different between high-fitness and low-fitness groups (p$<$0.01).}} 
\label{fig:heat}
\end{wrapfigure}

For VBM analysis, the voxels were compared individually to identify those which had significant differences in intensity across the tissue density maps. The heat maps illustrating voxelwise differences are identified in Figure \ref{fig:heat}. The clusters here are uncorrected for multiple comparisons, but significance was selected at level $p<0.01$. The clusters identified by VBM appear to be spatially distributed across the entire brain. There are changes identified both in occipital and frontal regions of gray matter, as well as in the periventricular white matter affecting frontal regions predominantly. Interestingly, these are some of the same regions identified to be affected by fitness in Figure \ref{fig:histogram}b. However, the global shifts in tissue profile such as atrophy are not captured or well-indexed by a voxelwise analysis, which is better suited for localizing to clusters. 

Therefore, while VBM is better suited to localize changes to specific clusters or anatomic regions, and DBM does not adequately assess tissue topology, transport-based morphometry is able to more fully assess the role of fitness on brain health on tissue distribution later in life to generate direct visualizations of the interface between the two classes. 

\subsection{Visualizing principal phenotypic variations using TBM for unsupervised learning}

\begin{figure}
\subfloat[]{\includegraphics[width=\linewidth]{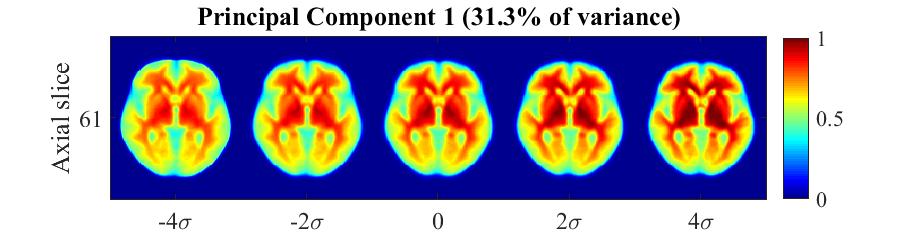}}
\newline
\subfloat[]{\includegraphics[width=\linewidth]{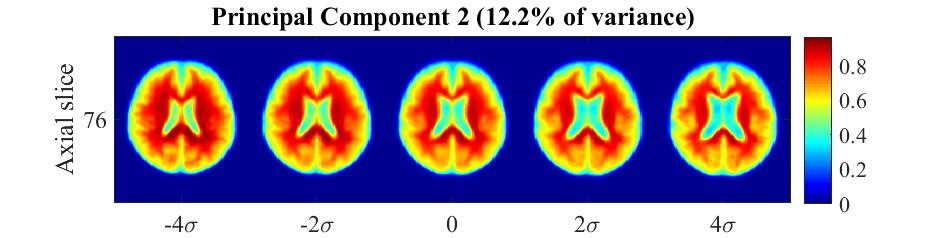}}
\newline
\subfloat[]{\includegraphics[width=\linewidth]{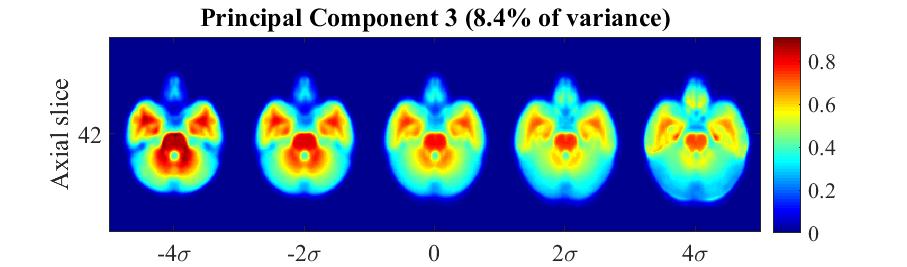}}
\caption{{\footnotesize{Visualizing top principal components in transport space. (a) PC1: variability in brain size, (b) PC2: variability in brain tissue atrophy and size of ventricles, (c) PC3: variability in prominence of midbrain and brainstem structures} }}
\label{fig:PCA}
\end{figure}

Finally, TBM can be used to visualize the top three PCA directions generated in transport space using TBM to gain a sense of the principal modes of variation in the dataset, which show variations in brain size, level of tissue atrophy, and prominence of midbrain structures shown by Figure \ref{fig:PCA}.

\section{Discussion}
\label{sec:discussion}

We demonstrate a fully automated technique for MRI analysis that facilitates discovery of structural shifts associated with observable phenotypes called transport-based morphometry (TBM). The results confirm our hypothesis that designing a TBM framework suitable for analysis of radiology images can facilitate tasks of regression, discrimination, and signal separation in the transform domain. Our approach is able to assess and visualize aging-related morphologic changes in a fully automated manner. The changes discovered independently by TBM match those that are well-accepted clinically. Additionally, our technique is able to investigate morphological differences between high-fitness and low-fitness groups and yield physically meaningful visualization of the interface between the two classes, suggesting a new mechanism by which fitness may mediate brain health later in life. Finally, transport-based morphometry is shown to enable signal separation by allowing visualization of biologically interpretable principal phenotypic variations. 


Traditional methods for assessing structural correlates in neuroimages, such as those that utilize numerical descriptors or pixelwise comparison are able to test only a subset of the information available. Deformation-based morphometry computes local volume expansion/contraction in terms of deformation fields, but cannot quantify differences in texture. We see that DBM is able to identify volume expansions, but the deformation fields lose information about tissue topology, distorting the texture of normal landmarks in the image. Voxel-based morphometry identifies voxels on a fixed grid, but cannot assess nonlinear or spatially diffuse changes such as atrophy or tissue thinning. Thus, results obtained using deformation-based analysis are influenced by limitations of the method, which confound biological insights. 

Transport-based morphometry analyzes tissue spatial distribution in the transform domain, where we see that even linear regression and discrimination techniques in the transform domain are sufficient to assess and visualize changes in tissue distribution that are nonlinear, spatially diffuse, and affect various regions of the brain in unequal ways. There are several reasons why transforming images to transport space enhances a range of pattern analysis tasks. First, optimal mass transport provides a metric by which to compare nonlinear signals whereby distances between images in the image domain can be modeled in terms of geodesics on a Riemannian manifold in transport space \cite{kolouri2016continuous, wang2013linear}. By projecting these geodesics locally to the tangent space, linearized versions of these metrics are available and as we see in this paper, Euclidean models in the transport space can capture a range of nonlinear morphologic changes. Second, the optimal mass preserving mapping has a unique minimizer with a bijective relationship with the source image with respect to a template. Therefore, in addition to enabling complex relationships to be more easily modeled, transport-based morphometry is a generative technique. An observable datapoint can be generated from any arbitrary point in the manifold. Because TBM is generative, a linear model can be directly visualized as a series of physically interpretable images through inverse TBM transformation. In contrast, VBM and DBM are not generative methods. We see indeed that compared to DBM and VBM, transport-based morphometry produces more direct visualizations. Using the TBM technique, we are able to adequately assess both global atrophy and local frontotemporal thinning with aging. In contrast, DBM was able to depict only local volume contraction and VBM was able to localize to individual voxels undergoing density changes. Furthermore, compared to VBM and DBM, transport-based morphometry coupled with discriminant analysis revealed a possible mechanism by which aerobic fitness mediates brain health later in life. By analyzing tissue spatial distribution using OMT, TBM can capture important phenomena that is not considered by VBM or DBM.

There are several limitations of this work. First, our approach for optimal transport minimization is non-convex. Although the approach does not guarantee theoretically that global minima will be achieved, the experimental results demonstrate that the multiscale scheme guides the minimization to the global minima and the results are comparable to those using convex formulations in 2D. We pose it as a future problem to couple the TBM framework presented in this paper with solvers that can overcome limitations with large 3D images and at the same time are convex. Another limitation is that analyzing the spatial distribution of voxels requires a normalization of images. Thus, the TBM transform does not directly consider whether there are statistically significant differences in the sum of voxel intensities. However, the latter limitation is easily remedied, as the sum of voxel intensities can be included as a feature when statistical analyses are performed in the feature domain. 

Finally, our formulation and TBM solver are fully general to any image modality and encompasses a wide range of problems in regression, discrimination, and unsupervised learning. Thus, our approach opens the door to numerous research and clinical advances. 

\section{Conclusion}
\label{sec:conclusion}

In conclusion, we presented a novel image transformation framework for MRI data to losslessly facilitate discovery of trends as well as yield biologically interpretable visualization of the morphologic changes associated with a variety of clinical outcomes. We demonstrate that our fully automated approach facilitates regression, discrimination, and blind signal separation with significant advancement over currently used techniques. Our approach is able to to independently discover aging-related changes that are well-corroborated clinically and provide new insight into the effects of fitness on the brain, unlike traditional methods. The results validate that our approach can be used as a statistical learning tool in diseases for which gene-structure-behavior relationships are not well-known.


%

\section*{Appendix A: Derivation of Euler-Lagrange Equation}
Here we present the derivation of the Euler-Lagrange equation in  \eqref{eq:EL}. Starting from the objective function in Eq. \eqref{eq:proposed} we have
\begin{eqnarray}
M(f)&=& \frac{1}{2}\|det(Df) I_1(f)-I_0\|^2+\frac{\lambda}{2}\| \nabla\times f \|^2 \nonumber\\
&=& \underbrace{\frac{1}{2}\int_\Omega (det(Df(x)) I_1(f(x))-I_0(x))^2 dx}_{M_1(f)} \nonumber \\
&&+ \underbrace{\frac{\lambda}{2}\int_\Omega |\nabla\times f(x)|^2 dx}_{M_2(f)}
\end{eqnarray}
where the first term, $M_1(f)$, enforces $f$ to be mass preserving while the second term, $M_2(f)$, enforces $f$ to be curl free. Starting with the first term we can write the Euler-Lagrange equation as, 
\begin{eqnarray}
\frac{d M_1}{d f^i}=\frac{\partial \L_1}{\partial f^i} - \sum_{k=1}^n \frac{d}{d x^k}(\frac{\partial \L_1}{\partial f_{x^k}^i}),~~~ i=1,...,n
\end{eqnarray}
where we have, 
\begin{eqnarray}
\frac{\partial \L_1}{\partial f^i}=det(Df)\frac{\partial I_1(f)}{\partial f^i}I_{error}
\label{eq:11term}
\end{eqnarray}
Let $C$ be the cofactor matrix of $Df$. Then $det(Df)$ can be written as the sum of the cofactors of any columns or rows of $Df$,
\begin{eqnarray}
det(Df)&=&\sum_{i=1}^n f^i_{x^j}C_{i,j},~ \forall j\in\{1,...,n\} \nonumber\\
&=&\sum_{j=1}^n f^i_{x^j}C_{i,j},~ \forall i\in\{1,...,n\} 
\end{eqnarray}
Using the cofactor matrix, C, we can write, 
\begin{eqnarray}
\frac{\partial \L_1}{\partial f^i_{x^k}}= C_{i,k}I_1(f)I_{error}
\label{eq:12term}
\end{eqnarray}
And from Equations \eqref{eq:11term} and \eqref{eq:12term} we have, 
\begin{eqnarray}
\frac{d M_1}{d f^i}= I_{error}(det(Df)\frac{\partial I_1(f)}{\partial f^i}-\sum_{k=1}^n\frac{d }{d x^k} C_{i,k}I_1(f))
\end{eqnarray}
and writing the vector form of the above equation for all $i$ and using $C^T=adj(Df)$ we can write, 
\begin{eqnarray}
\frac{d M_1}{d f}= I_{error}(det(Df)\nabla I_1(f)-\nabla\cdot (adj(Df)I_1(f))).
\label{eq:M1}
\end{eqnarray}

For the second term, we have
\begin{eqnarray}
\frac{\partial \L_1}{\partial f^i}=0
\label{eq:21term}
\end{eqnarray}
Furthermore, assuming that $n=2,3$ and using the Levi-Civita symbol $\epsilon$ we can write the norm squared of the curl of $f$ as follows, 
\begin{eqnarray}
|\nabla \times f|^2= \sum_{p=1}^n (\sum_{l=1}^n \sum_{m=1}^n \epsilon^{plm} f^m_{x^l})^2
\end{eqnarray}
which leads to,
\begin{eqnarray}
\frac{\partial \L_2}{\partial f_{x^k}^i}= \lambda  \sum_{p=1}^n \epsilon^{pki} (\sum_{l=1}^n \sum_{m=1}^n \epsilon^{plm} f^m_{x^l}).
\label{eq:22term}
\end{eqnarray}
Therefore we have, 
\begin{eqnarray}
\frac{d M_2}{d f^i}&=& -\lambda  \sum_{k=1}^n \frac{d}{dx^k} (\sum_{p=1}^n \epsilon^{pki} (\sum_{l=1}^n \sum_{m=1}^n \epsilon^{plm} f^m_{x^l}))\nonumber\\
&=& -\lambda  \sum_{k=1}^n\sum_{p=1}^n \epsilon^{pki} \frac{d}{dx^k} (\sum_{l=1}^n \sum_{m=1}^n \epsilon^{plm} f^m_{x^l})\nonumber\\
&=& \lambda  \sum_{k=1}^n\sum_{p=1}^n \epsilon^{ikp} \frac{d}{dx^k} (\sum_{l=1}^n \sum_{m=1}^n \epsilon^{plm} f^m_{x^l})\nonumber\\
&=& \lambda(\nabla \times \nabla \times f)^i 
\label{eq:M2}
\end{eqnarray}
Finally, combining Equations \eqref{eq:M1} and \eqref{eq:M2} will lead to, 
\begin{eqnarray}
\frac{d M}{d f}= I_{error}(det(Df)\nabla I_1(f)-\nabla\cdot (adj(Df)I_1(f))) \nonumber\\+\lambda \nabla \times \nabla \times f 
\end{eqnarray}


\section*{Appendix B: Validating OMT registration on MRI Data}

\subsection{Optimal transport minimization}
\label{ssec:challenges}

The equations for analysis and synthesis can be solved in closed form only for 1D signals \cite{villani2008optimal}, but for higher-dimensional signals, they must be solved using optimization techniques. Many solvers have been described in the OMT literature, although special challenges including drift, artifact, and computational time/complexity arise in numerical OMT of large image sets that may exceed millions of voxels.  

For example, Haker \etal \cite{haker2004} solve for an initial MP map $f_0$ (not unique or optimal) through the Knothe-Rosenblatt rearrangement \cite{villani2008optimal, bonnotte2013knothe} and then progressively update the initial map using composition (with another MP map $s$ that satisfies $s_\#\sigma=\sigma$) so that it becomes curl free to signify optimality. As pointed out by Haber \etal \cite{haber2010efficient} and Rehman \etal \cite{ur20093d}, however, there exist two main shortcomings to the preceding numerical approaches. First, a robust method is needed to obtain an initial MP mapping, and the obtained initial map is often far from the optimal transport map. Second, and much more importantly, such methods update the transformation in a space which is tangential to the linearized MP constraint. Hence, for any finite step update used in the optimization, $f_0(s^k)$, the mapping deviates or \textit{drifts} from the set of mass preserving mappings $MP$. While the level of \textit{drift} may or may not be acceptable in practice for a 2D solution, the drifting is amplified for 3D images as demonstrated in Section \ref{sec:results}. The \textit{drift} phenomenon necessitates solution by alternative methods for 3D images. 

Convergent methods have also been proposed based on a fluid dynamics formulation of the problem \cite{benamou2000computational,maas2015generalized} or based on the solution of the Monge-Amp\`{e}re equation \cite{benamou2014numerical}, but these formulations come at the cost of an additional virtual time dimension, which is computationally expensive. Computational cost also becomes a challenge in approaches utilizing a system of linear equations that arise from the finite-difference implementation of the linearized Monge-Ampere equation \cite{saumier2015efficient,oberman2015efficient}.

Another family of solvers \cite{solomon2015convolutional,cuturi2013sinkhorn, chartrand2009gradient} are based on Kantorovich's formulation of the problem. In short, Kantorovich's formulation searches for the optimal transport plan $\pi$ defined on $\Omega\times\Omega$ with marginals $\mu$ and $\sigma$ that minimizes the following, 
\begin{eqnarray}
\min_{\pi\in\Pi(\mu,\sigma)} \int_{\Omega\times\Omega} c(x,y)d\pi(x,y)
\label{eq:kantorovich}
\end{eqnarray}
Here $\Pi(\mu,\sigma)$ is the set of all transport plans with marginals $\mu$ and $\sigma$. Chartrand \textit{et al} \cite{chartrand2009gradient} solves the dual problem to the Kantorovich formulation. Chartrand \textit{et al} obtain the optimal transport map through a gradient descent solution. The obtained transport map as pointed out in \cite{chartrand2009gradient} and shown in this paper see Figure \ref{fig:3Dmorphed} comes with the trade off of undesired \textit{artifacts}, especially when the images are not smooth. Thus, additional work is needed to overcome challenges related to quality of MP match with the Chartrand \etal \cite{chartrand2009gradient} approach. 

In summary, in order to test the hypothesis that transport-based morphometry can both extract discriminant information and produce visualization of differences on MRI data, an OMT solution is required that can overcome computational challenges for large 3D data.

We validated our OMT approach on healthy, adult brain images obtained from the IXI dataset, Biomedical Image Analysis Group at the Imperial College in London \cite{BrainDevelopment}. 

10 images were selected at random from Guy's Hospital in UK. Subjects were male and ranged from 41 to 86 years of age at the time of imaging (mean age 57.8 years, standard deviation 15.7 years). The images were T1-weighted images, obtained using a Philips Medical Systems Intera 1.5 T scanner, with the following imaging parameters. Repetition time = 9.813 ms, echo time = 4.603 ms, number of phase encoding steps = 192, echo train length = 0, reconstruction diameter = 240, flip angle = 8$^\circ$. The images are 128 x 128 x 128 matrix, with 1 $mm^3$ resolution.

\subsubsection{Optimization parameters}

The step size for accelerated gradient descent is chosen such that the maximum displacement per update is 0.01 of a pixel, the same for our method and comparison methods. At every step of gradient descent, there is a check to maintain that the mapping is diffeomorphic and the step size is reduced as necessary in order to ensure a diffeomorphic mapping. The parameters were obtained experimentally: $\lambda = 100$, $\gamma = 6.5 \times 10^{4}$ when the MSE reaches 25\% of the initial MSE to steer the solution towards a curl-free MP mapping, number of scales = 3. The multiscale approach was also implemented for the two other OMT methods in this paper, although the results did not significantly change when the scheme was used.

The termination criteria for all methods implemented is when MSE of the morphed source image relative to the template image reaches 0.55\%. When the drifting phenomena in the Haker method \cite{haker2004optimal} produces MSE 100x the initial MSE, we terminate the code. We report the mean $L_2$ norm of the curl per voxel and MSE relative to the template.

We use a numerical discretization scheme in which values are placed at pixel or voxel centers. A consistent second-order finite difference approximation was used for all differential operators, utilizing the DGradient toolbox for MATLAB \cite{dgradient}.

\subsection{Experiment: Validating OMT registration on MRI}

We compare the approach of Haker \textit{et al} \cite{haker2004optimal} and Chartrand \textit{et al} \cite{chartrand2009gradient} to the approach described in this paper. The computational complexity of the gradient descent update step for all three methods implemented is $\O(NlogN)$. All methods implemented utilized the same preprocessed images. 

All unique pairs of images were registered to each other for the 10 images, resulting in 45 total registration problems. The statistics reported in this paper are based on the registrations performed in turn with our method, that of Chartrand \textit{et al}, and that of Haker \textit{et al}. 

The following three experiments investigate TBM for MRI pattern analysis enabled by our OMT approach.

\subsection{Comparing MP registration methods}

We report results for both 2D and 3D MP registration using optimal transport. The 2D image dataset was derived from the 3D dataset by extracting the same axial slice from the middle of every 3D brain image. The solver and accelerated gradient descent update equations are the same whether working with 2D or 3D images. We compare our OMT approach to current methods in the literature based on both the Monge formulation (\cite{haker2004optimal}) and Kantorovich formulation (\cite{chartrand2009gradient}) to demonstrate that our method is robust to the challenges of other OMT approaches. 
 
\subsubsection{2D optimal mass transport}

Table \ref{tab:example} displays the mapping statistics for the set of 2D images. The mass transported is lowest for the Chartrand \etal's method and compares favorably to the mass transport achieved using our method. The MSE for Chartrand \textit{et al} indicates that this method also produces the poorest MP mapping of the three methods. The MSE achieved is 3-4 times that achieved by the other two methods, demonstrating that this method is prone to artifacts. 

Our method achieves the lowest MSE in addition to mass transport distance. As reviewed in Section \ref{sec:review}, the optimal MP mapping is the MP mapping that achieves minimum mass transport. The curl, a measure of optimality of the MP mapping, is 0 by design for the method of Chartrand \etal as expected (see Section \ref{sec:review}). The mean curl is the highest for our method compared to the other two methods, but is still small in an absolute sense ($10^{-4}$). 

\begin{figure}
\center
\includegraphics[width=0.8\linewidth]
{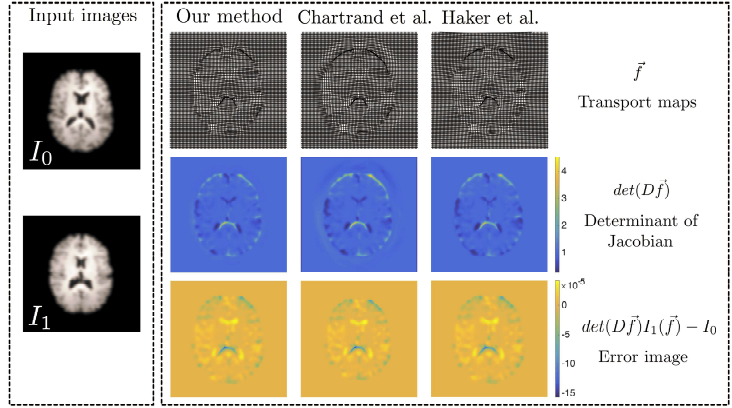}
\vspace{-0.5em}
\caption{\footnotesize{The source image $I_1$, target image $I_0$, and their calculated optimal transport map $\vec{f}$, corresponding determinant of Jacobian matrices, and the error image for the Haker method, the Chartrand method and our method. All are comparable for 2D OMT} }
\label{fig:Comparison}
\end{figure}

Our method produces the best results in terms of mass-preservation and mass transport. In Figure \ref{fig:Comparison}, we see the optimal transport fields and their corresponding morphings for several 2D brain images. Visually, the transport fields and quality of morphings are similar for all three methods. 

The proposed method was prototyped in MATLAB using built-in functions. The average runtime for $256\times 256$ brain images with $3$ scales was $ 19.36\pm 7.91$ seconds. 

\subsubsection{3D optimal mass transport}

\begin{figure*} 
\centering
\includegraphics[width = \textwidth]{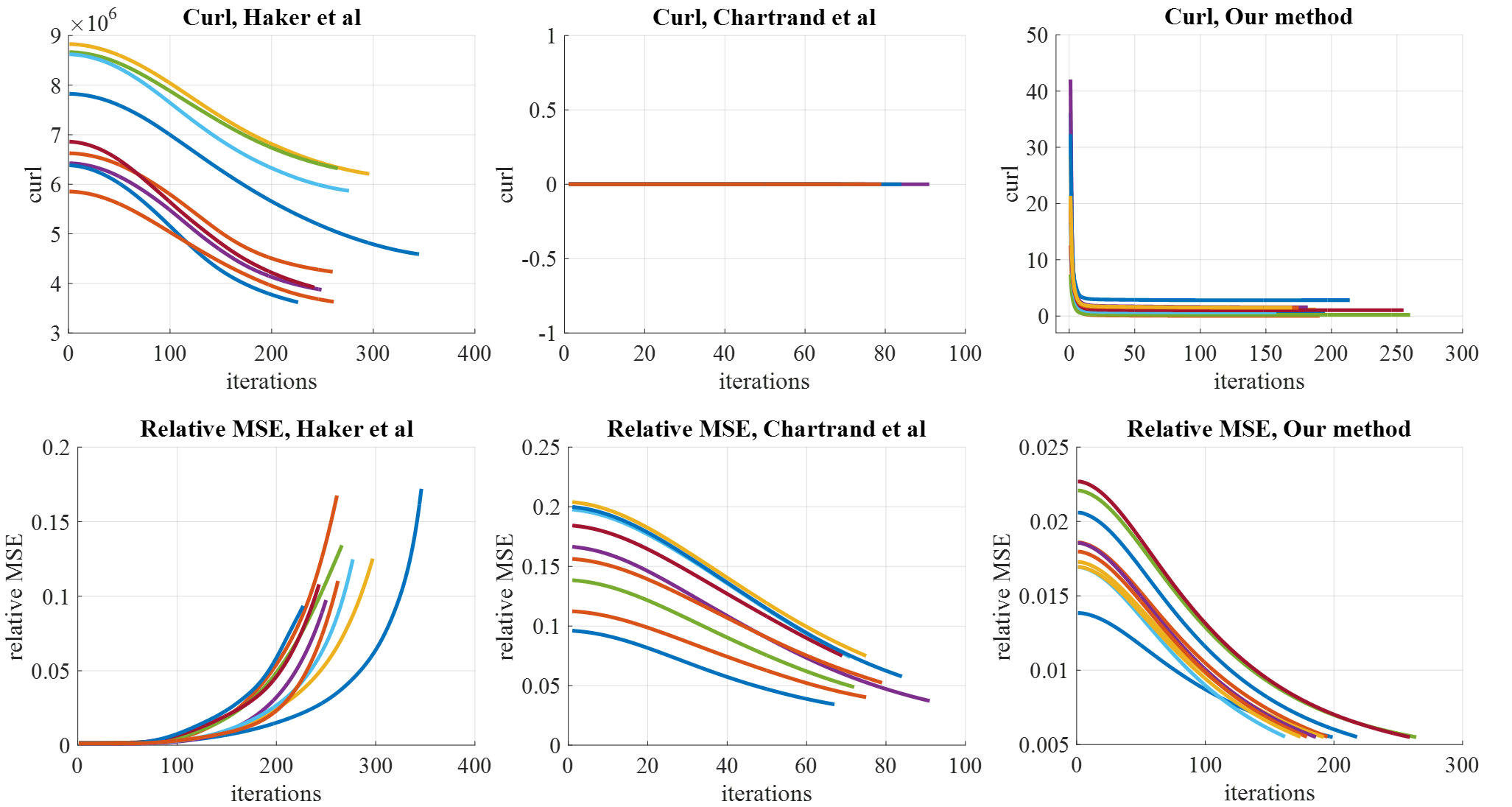}
\caption{\footnotesize{We see the plots for MSE, curl and mass transported for all three methods. The plots for our method are shown only for the last scale of the GP, using an initial point already close to the final point.}} 
\label{fig:3Dplots}
\end{figure*}

While in 2D all three methods seem to produce visually similar OMT mappings, we see in 3D that several phenomena become evident. Examining Figure \ref{fig:3Dplots}, we compare the plots of curl and relative MSE over gradient descent iterations for several brain images. We see that the magnitude of the curl (on the order of $10^6$) is large for the Haker \etal's method. The curl for the Chartrand \etal's method remains at 0 by design. Our method produces curls for all images that tend toward zero with iterations of gradient descent. 

\begin{figure}
\centering
\includegraphics[width= 0.8\columnwidth, angle = 270]{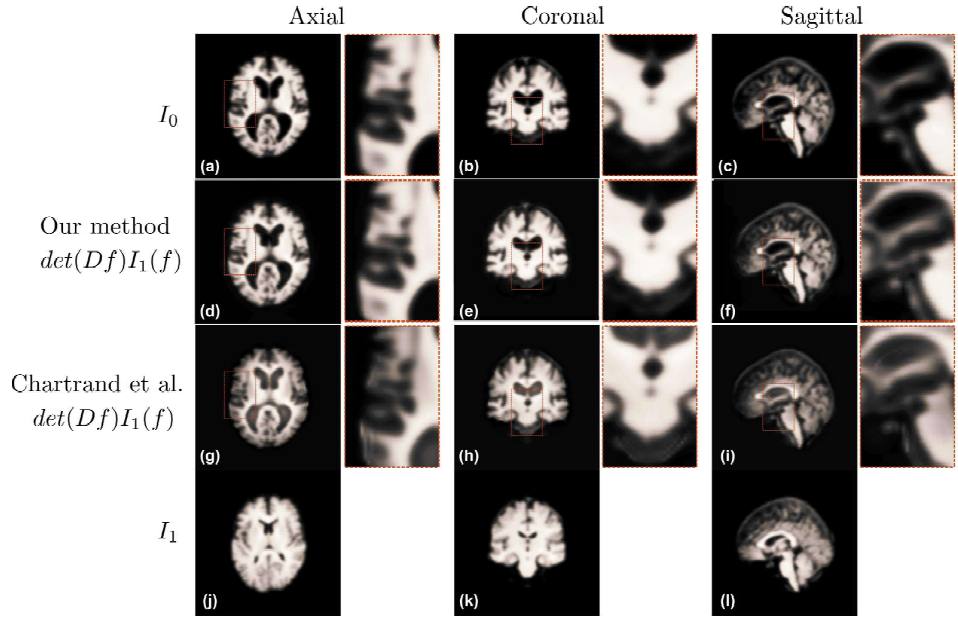}
\vspace{-0.5em}
\caption{\footnotesize{The target image (a-c), the morphed image in axial, coronal and sagittal cuts using our method (d-f) and the method presented by Chartrand \etal \cite{chartrand2009gradient} (g-i), and the source image (j-l)}}
\label{fig:3Dmorphed}
\end{figure}

We see that relative MSE with the Haker \etal's approach increases significantly until we terminate the code when the MSE reaches 100x its initial value. Hence, starting after around 100 iterations of gradient descent, the phenomenon of \textit{drift} with the Haker \etal's approach becomes evident. The relative MSE of the Chartrand \etal's approach decreases, but remains large in magnitude (5-10\%) at termination. The large MSE results in visual artifacts in the quality of the MP match, which we can see in Figure \ref{fig:3Dmorphed}. In contrast, for our method, all images are able to achieve the 0.55\% termination criterion. 

Table \ref{tab:example} corroborates the plots in Figure \ref{fig:3Dplots}. Our method produces the lowest relative MSE (best MP mapping), and all brain images are able to achieve the termination criterion of 0.55\%. Furthermore, our curl at termination is 8 orders of magnitude lower than that obtained using the Haker method. 

In terms of mass transported, the Chartrand \etal's method produces the lowest transport distance, although the MSE of the MP mapping is about 5-10x higher than that achieved using our method. We can also see artifacts visually in the mappings produced by the Chartrand \etal's method compared to our method (Figure \ref{fig:3Dmorphed}).

In Figure \ref{fig:3Dmorphed}, we compare axial, sagittal and coronal slices mapped using our method and that of Chartrand \etal's (The method of Haker \etal failed to produce a viable solution, which is why it is not shown.) We see that mappings produced by our method result in visually similar images to the target image $I_0$, whereas those produced by the Chartrand \etal's method contain several artifacts. 

Overall, our method outperforms both comparison methods for 3D images. Our method achieves the lowest MP mapping, while at the same time achieving small curl and mass transported.  

The median runtime was under 20 minutes per brain in MATLAB using built-in libraries on a general purpose computer. There is significant opportunity for improvement with an implementation in native C.    

Thus, we see that our approach is able to overcome traditional limitations of drift, artifact, and impractical computational complexity. Our approach enables the goal of pattern analysis on MRI using a transport-based morphometry approach.

\begin{table*} [t]
  \centering  
  \caption{Comparing methods of solving OMT in 2D}
  \begin{tabular}{|c|c|c|c|}
    \hline
    2D & \multicolumn{3}{c|}{\textbf{OT Mapping Statistics}}  \\ \hline
    \textbf{Method} & \emph{Relative MSE} & \emph{Mean curl} & \emph{Mass transported} \\ \hline
    Our method & $0.23 \pm 0.056 \%$ & $(8.7 \pm 5.3) \times 10^{-4}$ & $1.63\pm 0.57$ \\\hline
    Chartrand \etal & $1.8 \pm 2.9 \% $ & 0 & $1.56\pm 0.46$ \\\hline
   Haker \etal & $0.45 \pm 0.59 \%$ & $(7.0 \pm 0.16)\times 10^{-6} $ & $ 2.37 \pm 0.79$ \\\hline
  \end{tabular}%
  \end{table*}
  
  \begin{table*}
  \centering
  \caption{Comparing methods of solving OMT in 3D}
\begin{tabular}{|c|c|c|c|}
    \hline
    3D & \multicolumn{3}{c|}{\textbf{OT Mapping Statistics}}  \\ \hline
    \textbf{Method} & \emph{Relative MSE} & \emph{Mean curl}  & \emph{Mass transported}\\ \hline
    Our method & $0.55 \pm 0.0011 \%$ & $0.37 \pm 0.61$  & $1.3 \pm 0.50$ \\\hline
    Chartrand \etal & $3.0 \pm 1.8 \% $ & 0 & $0.07 \pm 0.04$\\\hline
    {Haker \etal} & {$9.9 \pm 2.5 \%$} & {$(4.5 \pm 1.6) \times 10^6$} & $12.5 \pm 4.8$\\\hline
  \end{tabular}
  \begin{flushright}
  \end{flushright}
  \label{tab:example}
\end{table*} 

\section*{Acknowledgment}

This work was supported in part by NSF award CCF 1421502, and NIH awards R01 GM090033 as well as the National Institute on Aging awards R01 AG25667 and R01 AG25302. This material is also based upon work supported by the Dowd-ICES graduate fellowship. The authors would like to thank Shlomo Ta'asan, Misha Lavrov for stimulating conversations.

\section*{References}

\bibliography{MultiscaleVariationalOT}

\begin{thebibliography}{41}
\expandafter\ifx\csname natexlab\endcsname\relax\def\natexlab#1{#1}\fi
\providecommand{\url}[1]{\texttt{#1}}
\providecommand{\href}[2]{#2}
\providecommand{\path}[1]{#1}
\providecommand{\DOIprefix}{doi:}
\providecommand{\ArXivprefix}{arXiv:}
\providecommand{\URLprefix}{URL: }
\providecommand{\Pubmedprefix}{pmid:}
\providecommand{\doi}[1]{\href{http://dx.doi.org/#1}{\path{#1}}}
\providecommand{\Pubmed}[1]{\href{pmid:#1}{\path{#1}}}
\providecommand{\bibinfo}[2]{#2}
\ifx\xfnm\relax \def\xfnm[#1]{\unskip,\space#1}\fi
\bibitem[{Ashburner(2007)}]{ashburner2007fast}
\bibinfo{author}{Ashburner, J.} (\bibinfo{year}{2007}).
\newblock \bibinfo{title}{A fast diffeomorphic image registration algorithm}.
\newblock {\it \bibinfo{journal}{Neuroimage}\/},  {\it \bibinfo{volume}{38}\/},
  \bibinfo{pages}{95--113}.
\bibitem[{Ashburner \& Friston(2000)}]{ashburner2000voxel}
\bibinfo{author}{Ashburner, J.}, \& \bibinfo{author}{Friston, K.~J.}
  (\bibinfo{year}{2000}).
\newblock \bibinfo{title}{Voxel-based morphometry the methods}.
\newblock {\it \bibinfo{journal}{Neuroimage}\/},  {\it \bibinfo{volume}{11}\/},
  \bibinfo{pages}{805--821}.
\bibitem[{Ashburner et~al.(2000)Ashburner, Good \&
  Friston}]{ashburner2000tensor}
\bibinfo{author}{Ashburner, J.}, \bibinfo{author}{Good, C.}, \&
  \bibinfo{author}{Friston, K.~J.} (\bibinfo{year}{2000}).
\newblock \bibinfo{title}{Tensor based morphometry}.
\newblock {\it \bibinfo{journal}{NeuroImage}\/},  {\it \bibinfo{volume}{11}\/},
  \bibinfo{pages}{S465}.
\bibitem[{Ashburner et~al.(1998)Ashburner, Hutton, Frackowiak, Johnsrude,
  Price, Friston et~al.}]{ashburner1998identifying}
\bibinfo{author}{Ashburner, J.}, \bibinfo{author}{Hutton, C.},
  \bibinfo{author}{Frackowiak, R.}, \bibinfo{author}{Johnsrude, I.},
  \bibinfo{author}{Price, C.}, \bibinfo{author}{Friston, K.} et~al.
  (\bibinfo{year}{1998}).
\newblock \bibinfo{title}{Identifying global anatomical differences:
  deformation-based morphometry}.
\newblock {\it \bibinfo{journal}{Human brain mapping}\/},  {\it
  \bibinfo{volume}{6}\/}, \bibinfo{pages}{348--357}.
\bibitem[{Basu et~al.(2014)Basu, Kolouri \& Rohde}]{basu2014detecting}
\bibinfo{author}{Basu, S.}, \bibinfo{author}{Kolouri, S.}, \&
  \bibinfo{author}{Rohde, G.~K.} (\bibinfo{year}{2014}).
\newblock \bibinfo{title}{Detecting and visualizing cell phenotype differences
  from microscopy images using transport-based morphometry}.
\newblock {\it \bibinfo{journal}{Proceedings of the National Academy of
  Sciences}\/},  {\it \bibinfo{volume}{111}\/}, \bibinfo{pages}{3448--3453}.
\bibitem[{Benamou \& Brenier(2000)}]{benamou2000computational}
\bibinfo{author}{Benamou, J.-D.}, \& \bibinfo{author}{Brenier, Y.}
  (\bibinfo{year}{2000}).
\newblock \bibinfo{title}{A computational fluid mechanics solution to the
  {M}onge-{K}antorovich mass transfer problem}.
\newblock {\it \bibinfo{journal}{Numerische Mathematik}\/},  {\it
  \bibinfo{volume}{84}\/}, \bibinfo{pages}{375--393}.
\bibitem[{Benamou et~al.(2014)Benamou, Froese \&
  Oberman}]{benamou2014numerical}
\bibinfo{author}{Benamou, J.-D.}, \bibinfo{author}{Froese, B.~D.}, \&
  \bibinfo{author}{Oberman, A.~M.} (\bibinfo{year}{2014}).
\newblock \bibinfo{title}{Numerical solution of the optimal transportation
  problem using the monge--ampere equation}.
\newblock {\it \bibinfo{journal}{Journal of Computational Physics}\/},  {\it
  \bibinfo{volume}{260}\/}, \bibinfo{pages}{107--126}.
\bibitem[{Bonnotte(2013)}]{bonnotte2013knothe}
\bibinfo{author}{Bonnotte, N.} (\bibinfo{year}{2013}).
\newblock \bibinfo{title}{From {K}nothe's rearrangement to {B}renier's optimal
  transport map}.
\newblock {\it \bibinfo{journal}{SIAM Journal on Mathematical Analysis}\/},
  {\it \bibinfo{volume}{45}\/}, \bibinfo{pages}{64--87}.
\bibitem[{Bookstein(2001)}]{bookstein2001voxel}
\bibinfo{author}{Bookstein, F.~L.} (\bibinfo{year}{2001}).
\newblock \bibinfo{title}{``{V}oxel-based morphometry'' should not be used with
  imperfectly registered images}.
\newblock {\it \bibinfo{journal}{Neuroimage}\/},  {\it \bibinfo{volume}{14}\/},
  \bibinfo{pages}{1454--1462}.
\bibitem[{Brenier(1991)}]{brenier1991}
\bibinfo{author}{Brenier, Y.} (\bibinfo{year}{1991}).
\newblock \bibinfo{title}{Polar factorization and monotone rearrangement of
  vector-valued functions}.
\newblock {\it \bibinfo{journal}{Communications on pure and applied
  mathematics}\/},  {\it \bibinfo{volume}{44}\/}, \bibinfo{pages}{375--417}.
\bibitem[{Brody(1970)}]{brody1970structural}
\bibinfo{author}{Brody, H.} (\bibinfo{year}{1970}).
\newblock \bibinfo{title}{Structural changes in the aging nervous system}.
\newblock In {\it \bibinfo{booktitle}{The regulatory role of the nervous system
  in aging}\/} (pp. \bibinfo{pages}{9--21}).
\newblock \bibinfo{publisher}{Karger Publishers}.
\bibitem[{Carlson(2011)}]{Carlson2011}
\bibinfo{author}{Carlson, N.} (\bibinfo{year}{2011}).
\newblock \bibinfo{title}{Fluorescence guided resection}.
\newblock \bibinfo{note}{\url{https://engineering.dartmouth.edu/brainidb}}.
\bibitem[{Chartrand et~al.(2009)Chartrand, Vixie, Wohlberg \&
  Bollt}]{chartrand2009gradient}
\bibinfo{author}{Chartrand, R.}, \bibinfo{author}{Vixie, K.},
  \bibinfo{author}{Wohlberg, B.}, \& \bibinfo{author}{Bollt, E.}
  (\bibinfo{year}{2009}).
\newblock \bibinfo{title}{A gradient descent solution to the
  {M}onge-{K}antorovich problem}.
\newblock {\it \bibinfo{journal}{Applied Mathematical Sciences}\/},  {\it
  \bibinfo{volume}{3}\/}, \bibinfo{pages}{1071--1080}.
\bibitem[{Cootes et~al.(1995)Cootes, Taylor, Cooper \&
  Graham}]{cootes1995active}
\bibinfo{author}{Cootes, T.~F.}, \bibinfo{author}{Taylor, C.~J.},
  \bibinfo{author}{Cooper, D.~H.}, \& \bibinfo{author}{Graham, J.}
  (\bibinfo{year}{1995}).
\newblock \bibinfo{title}{Active shape models-their training and application}.
\newblock {\it \bibinfo{journal}{Computer vision and image understanding}\/},
  {\it \bibinfo{volume}{61}\/}, \bibinfo{pages}{38--59}.
\bibitem[{Cuturi(2013)}]{cuturi2013sinkhorn}
\bibinfo{author}{Cuturi, M.} (\bibinfo{year}{2013}).
\newblock \bibinfo{title}{Sinkhorn distances: Lightspeed computation of optimal
  transport}.
\newblock In {\it \bibinfo{booktitle}{Advances in Neural Information Processing
  Systems}\/} (pp. \bibinfo{pages}{2292--2300}).
\bibitem[{Erickson et~al.(2009)Erickson, Prakash, Voss, Chaddock, Hu, Morris,
  White, W{\'o}jcicki, McAuley \& Kramer}]{erickson2009aerobic}
\bibinfo{author}{Erickson, K.~I.}, \bibinfo{author}{Prakash, R.~S.},
  \bibinfo{author}{Voss, M.~W.}, \bibinfo{author}{Chaddock, L.},
  \bibinfo{author}{Hu, L.}, \bibinfo{author}{Morris, K.~S.},
  \bibinfo{author}{White, S.~M.}, \bibinfo{author}{W{\'o}jcicki, T.~R.},
  \bibinfo{author}{McAuley, E.}, \& \bibinfo{author}{Kramer, A.~F.}
  (\bibinfo{year}{2009}).
\newblock \bibinfo{title}{Aerobic fitness is associated with hippocampal volume
  in elderly humans}.
\newblock {\it \bibinfo{journal}{Hippocampus}\/},  {\it
  \bibinfo{volume}{19}\/}, \bibinfo{pages}{1030--1039}.
\bibitem[{Fischl \& Dale(2000)}]{fischl2000measuring}
\bibinfo{author}{Fischl, B.}, \& \bibinfo{author}{Dale, A.~M.}
  (\bibinfo{year}{2000}).
\newblock \bibinfo{title}{Measuring the thickness of the human cerebral cortex
  from magnetic resonance images}.
\newblock {\it \bibinfo{journal}{Proceedings of the National Academy of
  Sciences}\/},  {\it \bibinfo{volume}{97}\/}, \bibinfo{pages}{11050--11055}.
\bibitem[{Friston \& Ashburner(2004)}]{friston2004generative}
\bibinfo{author}{Friston, K.}, \& \bibinfo{author}{Ashburner, J.}
  (\bibinfo{year}{2004}).
\newblock \bibinfo{title}{Generative and recognition models for neuroanatomy}.
\newblock {\it \bibinfo{journal}{Neuroimage}\/},  {\it \bibinfo{volume}{23}\/},
  \bibinfo{pages}{21--24}.
\bibitem[{Haber et~al.(2010)Haber, Rehman \& Tannenbaum}]{haber2010efficient}
\bibinfo{author}{Haber, E.}, \bibinfo{author}{Rehman, T.}, \&
  \bibinfo{author}{Tannenbaum, A.} (\bibinfo{year}{2010}).
\newblock \bibinfo{title}{An efficient numerical method for the solution of the
  {L}\_2 optimal mass transfer problem}.
\newblock {\it \bibinfo{journal}{SIAM Journal on Scientific Computing}\/},
  {\it \bibinfo{volume}{32}\/}, \bibinfo{pages}{197--211}.
\bibitem[{Haker et~al.(2004{\natexlab{a}})Haker, Zhu, Tannenbaum \&
  Angenent}]{haker2004}
\bibinfo{author}{Haker, S.}, \bibinfo{author}{Zhu, L.},
  \bibinfo{author}{Tannenbaum, A.}, \& \bibinfo{author}{Angenent, S.}
  (\bibinfo{year}{2004}{\natexlab{a}}).
\newblock \bibinfo{title}{Optimal mass transport for registration and warping}.
\newblock {\it \bibinfo{journal}{International Journal of Computer Vision}\/},
  {\it \bibinfo{volume}{60}\/}, \bibinfo{pages}{225--240}.
\bibitem[{Haker et~al.(2004{\natexlab{b}})Haker, Zhu, Tannenbaum \&
  Angenent}]{haker2004optimal}
\bibinfo{author}{Haker, S.}, \bibinfo{author}{Zhu, L.},
  \bibinfo{author}{Tannenbaum, A.}, \& \bibinfo{author}{Angenent, S.}
  (\bibinfo{year}{2004}{\natexlab{b}}).
\newblock \bibinfo{title}{Optimal mass transport for registration and warping}.
\newblock {\it \bibinfo{journal}{International Journal of Computer Vision}\/},
  {\it \bibinfo{volume}{60}\/}, \bibinfo{pages}{225--240}.
\bibitem[{Imperial College London()}]{BrainDevelopment}
Imperial College London (\bibinfo{year}{accessed 3-21-16}).
\newblock
  \bibinfo{howpublished}{\url{http://brain-development.org/ixi-dataset/}}.
\bibitem[{Jeffery(2002)}]{jeffery2002use}
\bibinfo{author}{Jeffery, D.~R.} (\bibinfo{year}{2002}).
\newblock \bibinfo{title}{The use of vaccinations in patients with multiple
  sclerosis}.
\newblock {\it \bibinfo{journal}{Infections in medicine}\/},  {\it
  \bibinfo{volume}{19}\/}, \bibinfo{pages}{73--79}.
\bibitem[{Kolouri et~al.(2016{\natexlab{a}})Kolouri, Park \&
  Rohde}]{kolouri2016radon}
\bibinfo{author}{Kolouri, S.}, \bibinfo{author}{Park, S.~R.}, \&
  \bibinfo{author}{Rohde, G.~K.} (\bibinfo{year}{2016}{\natexlab{a}}).
\newblock \bibinfo{title}{The radon cumulative distribution transform and its
  application to image classification}.
\newblock {\it \bibinfo{journal}{IEEE transactions on image processing}\/},
  {\it \bibinfo{volume}{25}\/}, \bibinfo{pages}{920--934}.
\bibitem[{Kolouri et~al.(2016{\natexlab{b}})Kolouri, Tosun, Ozolek \&
  Rohde}]{kolouri2016continuous}
\bibinfo{author}{Kolouri, S.}, \bibinfo{author}{Tosun, A.~B.},
  \bibinfo{author}{Ozolek, J.~A.}, \& \bibinfo{author}{Rohde, G.~K.}
  (\bibinfo{year}{2016}{\natexlab{b}}).
\newblock \bibinfo{title}{A continuous linear optimal transport approach for
  pattern analysis in image datasets}.
\newblock {\it \bibinfo{journal}{Pattern recognition}\/},  {\it
  \bibinfo{volume}{51}\/}, \bibinfo{pages}{453--462}.
\bibitem[{Maas et~al.(2015)Maas, Rumpf, Sch{\"o}nlieb \&
  Simon}]{maas2015generalized}
\bibinfo{author}{Maas, J.}, \bibinfo{author}{Rumpf, M.},
  \bibinfo{author}{Sch{\"o}nlieb, C.}, \& \bibinfo{author}{Simon, S.}
  (\bibinfo{year}{2015}).
\newblock \bibinfo{title}{A generalized model for optimal transport of images
  including dissipation and density modulation}.
\newblock {\it \bibinfo{journal}{arXiv preprint arXiv:1504.01988}\/}, .
\bibitem[{Mathworks File Exchange()}]{dgradient}
Mathworks File Exchange (\bibinfo{year}{2016}).
\newblock \bibinfo{title}{Dgradient}.
\newblock
  \bibinfo{howpublished}{\url{http://www.mathworks.com/matlabcentral/fileexchange/29887-dgradient}}.
\bibitem[{Mechelli et~al.(2005)Mechelli, Price, Friston \&
  Ashburner}]{mechelli2005voxel}
\bibinfo{author}{Mechelli, A.}, \bibinfo{author}{Price, C.~J.},
  \bibinfo{author}{Friston, K.~J.}, \& \bibinfo{author}{Ashburner, J.}
  (\bibinfo{year}{2005}).
\newblock \bibinfo{title}{Voxel-based morphometry of the human brain: methods
  and applications}.
\newblock {\it \bibinfo{journal}{Current medical imaging reviews}\/},  {\it
  \bibinfo{volume}{1}\/}, \bibinfo{pages}{105--113}.
\bibitem[{Nesterov et~al.(2007)}]{nesterov2007gradient}
\bibinfo{author}{Nesterov, Y.} et~al. (\bibinfo{year}{2007}).
\newblock {\it \bibinfo{title}{Gradient methods for minimizing composite
  objective function}\/}.
\newblock \bibinfo{type}{Technical Report} UCL.
\bibitem[{for Neuroimaging(2016)}]{spm12}
\bibinfo{author}{for Neuroimaging, W. T.~C.} (\bibinfo{year}{2016}).
\newblock \bibinfo{title}{Spm12}.
\newblock \URLprefix \url{http://www.fil.ion.ucl.ac.uk/spm/software/spm12/}.
\bibitem[{Oberman \& Ruan(2015)}]{oberman2015efficient}
\bibinfo{author}{Oberman, A.~M.}, \& \bibinfo{author}{Ruan, Y.}
  (\bibinfo{year}{2015}).
\newblock \bibinfo{title}{An efficient linear programming method for optimal
  transportation}.
\newblock {\it \bibinfo{journal}{arXiv preprint arXiv:1509.03668}\/}, .
\bibitem[{Park et~al.(2017)Park, Kolouri, Kundu \& Rohde}]{park2017cumulative}
\bibinfo{author}{Park, S.~R.}, \bibinfo{author}{Kolouri, S.},
  \bibinfo{author}{Kundu, S.}, \& \bibinfo{author}{Rohde, G.~K.}
  (\bibinfo{year}{2017}).
\newblock \bibinfo{title}{The cumulative distribution transform and linear
  pattern classification}.
\newblock {\it \bibinfo{journal}{Applied and Computational Harmonic
  Analysis}\/}, .
\bibitem[{ur~Rehman et~al.(2009)ur~Rehman, Haber, Pryor, Melonakos \&
  Tannenbaum}]{ur20093d}
\bibinfo{author}{ur~Rehman, T.}, \bibinfo{author}{Haber, E.},
  \bibinfo{author}{Pryor, G.}, \bibinfo{author}{Melonakos, J.}, \&
  \bibinfo{author}{Tannenbaum, A.} (\bibinfo{year}{2009}).
\newblock \bibinfo{title}{3{D} nonrigid registration via optimal mass transport
  on the {GPU}}.
\newblock {\it \bibinfo{journal}{Medical image analysis}\/},  {\it
  \bibinfo{volume}{13}\/}, \bibinfo{pages}{931--940}.
\bibitem[{Salmond et~al.(2002)Salmond, Ashburner, Vargha-Khadem, Connelly,
  Gadian \& Friston}]{salmond2002distributional}
\bibinfo{author}{Salmond, C.}, \bibinfo{author}{Ashburner, J.},
  \bibinfo{author}{Vargha-Khadem, F.}, \bibinfo{author}{Connelly, A.},
  \bibinfo{author}{Gadian, D.}, \& \bibinfo{author}{Friston, K.}
  (\bibinfo{year}{2002}).
\newblock \bibinfo{title}{Distributional assumptions in voxel-based
  morphometry}.
\newblock {\it \bibinfo{journal}{Neuroimage}\/},  {\it \bibinfo{volume}{17}\/},
  \bibinfo{pages}{1027--1030}.
\bibitem[{Saumier et~al.(2015)Saumier, Agueh \&
  Khouider}]{saumier2015efficient}
\bibinfo{author}{Saumier, L.-P.}, \bibinfo{author}{Agueh, M.}, \&
  \bibinfo{author}{Khouider, B.} (\bibinfo{year}{2015}).
\newblock \bibinfo{title}{An efficient numerical algorithm for the {L}2 optimal
  transport problem with periodic densities}.
\newblock {\it \bibinfo{journal}{IMA Journal of Applied Mathematics}\/},  {\it
  \bibinfo{volume}{80}\/}, \bibinfo{pages}{135--157}.
\bibitem[{Shamir et~al.(2008)Shamir, Orlov, Eckley, Macura, Johnston \&
  Goldberg}]{shamir2008wndchrm}
\bibinfo{author}{Shamir, L.}, \bibinfo{author}{Orlov, N.},
  \bibinfo{author}{Eckley, D.~M.}, \bibinfo{author}{Macura, T.},
  \bibinfo{author}{Johnston, J.}, \& \bibinfo{author}{Goldberg, I.~G.}
  (\bibinfo{year}{2008}).
\newblock \bibinfo{title}{Wndchrm--an open source utility for biological image
  analysis}.
\newblock {\it \bibinfo{journal}{Source code for biology and medicine}\/},
  {\it \bibinfo{volume}{3}\/}, \bibinfo{pages}{1}.
\bibitem[{Solomon et~al.(2015)Solomon, de~Goes, Studios, Peyr{\'e}, Cuturi,
  Butscher, Nguyen, Du \& Guibas}]{solomon2015convolutional}
\bibinfo{author}{Solomon, J.}, \bibinfo{author}{de~Goes, F.},
  \bibinfo{author}{Studios, P.~A.}, \bibinfo{author}{Peyr{\'e}, G.},
  \bibinfo{author}{Cuturi, M.}, \bibinfo{author}{Butscher, A.},
  \bibinfo{author}{Nguyen, A.}, \bibinfo{author}{Du, T.}, \&
  \bibinfo{author}{Guibas, L.} (\bibinfo{year}{2015}).
\newblock \bibinfo{title}{Convolutional wasserstein distances: Efficient
  optimal transportation on geometric domains}.
\newblock {\it \bibinfo{journal}{ACM Transactions on Graphics (Proc. SIGGRAPH
  2015), to appear}\/}, .
\bibitem[{Villani(2008)}]{villani2008optimal}
\bibinfo{author}{Villani, C.} (\bibinfo{year}{2008}).
\newblock {\it \bibinfo{title}{Optimal transport: old and new}\/} volume
  \bibinfo{volume}{338}.
\newblock \bibinfo{publisher}{Springer Science \& Business Media}.
\bibitem[{Wang et~al.(2011{\natexlab{a}})Wang, Mo, Ozolek \&
  Rohde}]{wang2011penalized}
\bibinfo{author}{Wang, W.}, \bibinfo{author}{Mo, Y.}, \bibinfo{author}{Ozolek,
  J.~A.}, \& \bibinfo{author}{Rohde, G.~K.}
  (\bibinfo{year}{2011}{\natexlab{a}}).
\newblock \bibinfo{title}{Penalized fisher discriminant analysis and its
  application to image-based morphometry}.
\newblock {\it \bibinfo{journal}{Pattern recognition letters}\/},  {\it
  \bibinfo{volume}{32}\/}, \bibinfo{pages}{2128--2135}.
\bibitem[{Wang et~al.(2011{\natexlab{b}})Wang, Ozolek, Slepcev, Lee, Chen \&
  Rohde}]{wang2011optimal}
\bibinfo{author}{Wang, W.}, \bibinfo{author}{Ozolek, J.~A.},
  \bibinfo{author}{Slepcev, D.}, \bibinfo{author}{Lee, A.~B.},
  \bibinfo{author}{Chen, C.}, \& \bibinfo{author}{Rohde, G.~K.}
  (\bibinfo{year}{2011}{\natexlab{b}}).
\newblock \bibinfo{title}{An optimal transportation approach for nuclear
  structure-based pathology}.
\newblock {\it \bibinfo{journal}{Medical Imaging, IEEE Transactions on}\/},
  {\it \bibinfo{volume}{30}\/}, \bibinfo{pages}{621--631}.
\bibitem[{Wang et~al.(2013)Wang, Slepcev, Basu, Ozolek \&
  Rohde}]{wang2013linear}
\bibinfo{author}{Wang, W.}, \bibinfo{author}{Slepcev, D.},
  \bibinfo{author}{Basu, S.}, \bibinfo{author}{Ozolek, J.~A.}, \&
  \bibinfo{author}{Rohde, G.~K.} (\bibinfo{year}{2013}).
\newblock \bibinfo{title}{A linear optimal transportation framework for
  quantifying and visualizing variations in sets of images}.
\newblock {\it \bibinfo{journal}{International journal of computer vision}\/},
  {\it \bibinfo{volume}{101}\/}, \bibinfo{pages}{254--269}.

\end{thebibliography}

\section*{Vitae}

\textbf{Shinjini Kundu} received her B.S. and M.S. degrees in electrical engineering from Stanford University and her Ph.D. in biomedical engineering from Carnegie Mellon University. She is part of the Medical Scientist Training Program (MSTP) at the University of Pittsburgh, where she is currently pursuing her M.D. degree. Her research interests are in pattern recognition for biomedical images, magnetic resonance imaging, and computer-aided detection for radiology.

\textbf{Soheil Kolouri} received his B.S. degree in electrical engineering from Sharif University of Technology, Tehran, Iran, in 2010, and his M.S. degree in electrical engineering in 2012 from Colorado State University, Fort Collins, Colorado. He received his doctorate degree in biomedical engineering from Carnegie Mellon University in 2015, were his research was focused on applications of the optimal transport in image modeling, computer vision, and pattern recognition. His thesis, titled, “Transport-based pattern recognition and image modeling”, won the best thesis award from the Biomedical Engineering Department at Carnegie Mellon University.

\textbf{Kirk Erickson} received his B.S. degree in Psychology and Philosophy in 1999 from Marquette University. In 2005 he received his Ph.D. from the University of Illinois at Urbana-Champaign and was a post-doc at the Beckman Institute for Advanced Science and Technology at the University of Illinois until 2008. He is currently an Associate Professor of Psychology and the Center for the Neural Basis of Cognition at the University of Pittsburgh.

\textbf{Arthur Kramer} is the Director of the Beckman Institute for Advanced Science \& Technology and the Swanlund Chair and Professor of Psychology and Neuroscience at the University of Illinois. A major focus of his labs recent research is the understanding and enhancement of cognitive and neural plasticity across the lifespan. Professor Kramer is a fellow of the American Psychological Association, American Psychological Society, and a recipient of a NIH Ten Year MERIT Award. Professor Kramers research has been featured in a long list of print, radio and electronic media including the New York Times, Wall Street Journal, Washington Post, Chicago Tribune, CBS Evening News, Today Show, NPR and Saturday Night Live.

\textbf{Edward McAuley} is a Shahid and Ann Carlson Khan Endowed Professor of Applied Health Sciences at the University of Illinois at Urbana-Champaign. He holds appointments in the Departments of Kinesiology and Community Health, Psychology, Internal Medicine, and the Beckman Institute for Advanced Science and Technology. He the director of the Exercise Psychology Laboratory at Illinois and has published over 380 articles and chapters. He has served as the Chair of the Psychosocial Risk and Disease Prevention study section of the National Institutes of Health. He is an elected fellow of the Society of Behavioral Medicine and the Gerontological Society of America. His research agenda has focused primarily on physical activity, aging, and well-being in healthy adults and breast cancer survivors and the role played by exercise training in neurocognitive function, brain health, and psychological well-being.

\textbf{Gustavo K. Rohde} earned B.S. degrees in physics and mathematics in 1999, and the M.S. degree in electrical engineering in 2001 from Vanderbilt University. He received a doctorate in applied mathematics and scientific computation in 2005 from the University of Maryland. He is currently an associate professor of Biomedical Engineering and Electrical and Computer Engineering at University of Virginia.

\end{document}